\documentclass{CUP-JNL-DCE}

\usepackage{graphicx}
\usepackage{lmodern}
\usepackage[T1]{fontenc}
\usepackage[colorlinks=true,citecolor=red,linkcolor=black]{hyperref}
\usepackage[figurename=Fig.,labelfont=bf,labelsep=period]{caption}
\usepackage{bm}
\usepackage{tabularx}
\usepackage{makecell}
\usepackage{multirow}
\usepackage[table,xcdraw]{xcolor}
\usepackage{array}
\usepackage{lineno}
\usepackage{geometry} 
\usepackage{floatrow}
\DeclareFloatFont{footnotesize}{\footnotesize}
\usepackage{latexsym}
\usepackage{graphicx}
\usepackage{multicol,multirow}
\usepackage{amsmath,amssymb,amsfonts}
\usepackage{mathrsfs}
\usepackage{amsthm}
\usepackage{apacite}
\usepackage{rotating}
\usepackage{appendix}
\usepackage[authoryear]{natbib}
\usepackage{ifpdf}
\usepackage[T1]{fontenc}
\usepackage{times}
\usepackage{sourcesanspro}
\usepackage{newtxmath}
\usepackage{textcomp}%
\usepackage{xcolor}%
\usepackage{hyperref}

\definecolor{red}{RGB}{204,0,0}
\definecolor{orange}{RGB}{204,102,0}
\definecolor{blue}{RGB}{0,0,204}

\begin{document}

\begin{Frontmatter}

\title[Article Title]{ARE LARGE PRE-TRAINED VISION LANGUAGE MODELS EFFECTIVE CONSTRUCTION SAFETY INSPECTORS}

\author[1]{Xuezheng Chen}\email{xuezheng@student.ubc.ca}
\author*[2]{Zhengbo Zou}\email{zhengbo.zou@columbia.edu}

\authormark{Chen and Zou}

\address[1]{\orgdiv{Mechanical Engineering}, \orgname{University of British Columbia}, \orgaddress{\street{6250 Applied Science Lane}, \state{Vancouver}, \postcode{BC, V6T 1Z4}, \country{Canada}}}

\address*[2]{\orgdiv{Civil Engineering and Engineering Mechanics}, \orgname{Columbia University}, \orgaddress{\street{610 SW Mudd}, \state{New York}, \postcode{NY, 10027}, \country{U.S.}}}

\abstract{Construction safety inspections typically involve a human inspector identifying safety concerns on-site. With the rise of powerful Vision Language Models (VLMs), researchers are exploring their use for tasks such as detecting safety rule violations from on-site images. However, there is a lack of open datasets to comprehensively evaluate and further fine-tune VLMs in construction safety inspection. Current applications of VLMs use small, supervised datasets, limiting their applicability in tasks they are not directly trained for. In this paper, we propose the \textit{ConstructionSite 10k}, featuring 10,000 construction site images with annotations for three inter-connected tasks, including image captioning, safety rule violation visual question answering (VQA), and construction element visual grounding. Our subsequent evaluation of current state-of-the-art large pre-trained VLMs shows notable generalization abilities in zero-shot and few-shot settings, while additional training is needed to make them applicable to actual construction sites. This dataset allows researchers to train and evaluate their own VLMs with new architectures and techniques, providing a valuable benchmark for construction safety inspection.}


\begin{policy}[Impact Statement]
Vision-Language Models (VLMs) are a type of AI model that processes images as input and generates textual output. Large pre-trained VLMs, equipped with billions of parameters and trained on vast image-text pairs from the internet, are capable of performing tasks they were never explicitly trained on, using only a few prompts. This flexibility allows such models to be applied to a wide range of construction safety-related tasks without requiring changes to their architecture or additional training. This paper introduces an evaluation framework for assessing how well large pre-trained VLMs can perform construction safety tasks, along with a comprehensive dataset specifically designed for this purpose. Practitioners can use this dataset to fine-tune their own VLM to suit specific needs for safety inspection. 
\end{policy}

\end{Frontmatter}


\section{Introduction}
\label{introduction}

\begin{figure*}[ht!]
    \centering
    \includegraphics[width=\textwidth]{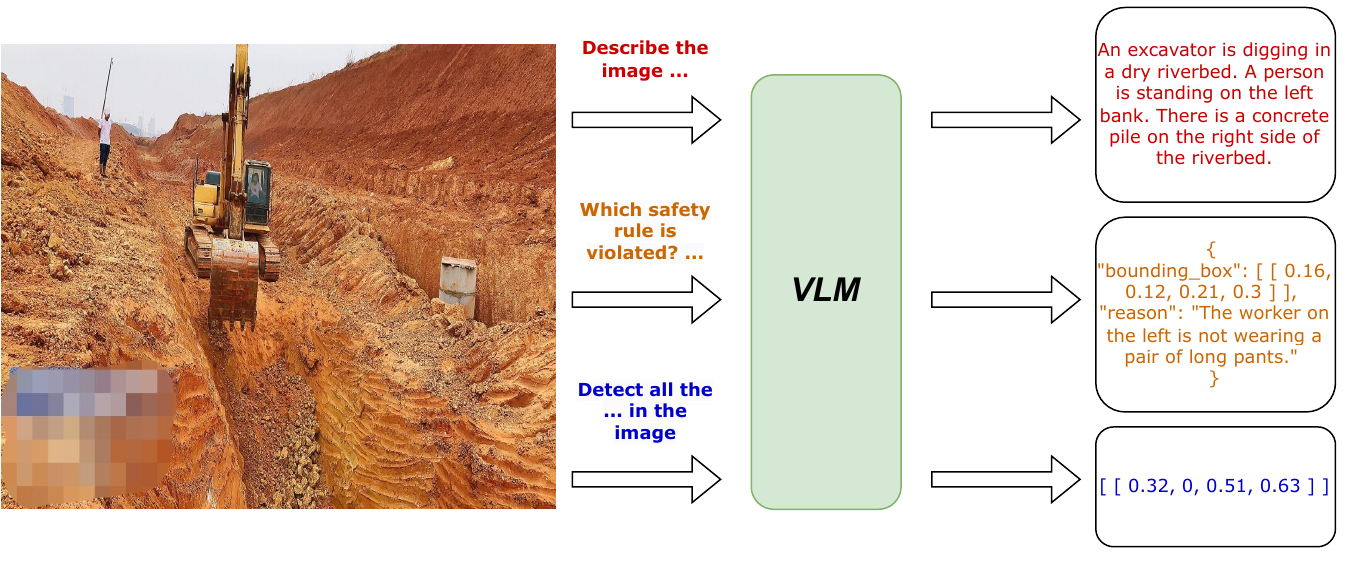}
    \caption{The high-level schematic diagram shows the three tasks employed in this paper. The VLM will receive the construction site image and prompts for three tasks: \textcolor{red}{image captioning}, \textcolor{orange}{safety rule violation VQA}, and \textcolor{blue}{construction element visual grounding}.}
    \label{fig:summary workflow}
\end{figure*}

Hazardous working conditions and risky behaviors at construction sites accounted for a concerning 21\% of all workplace fatalities in the U.S. in 2022 \citep{USbureauoflaborstatistics}. To ensure compliance with safety regulations, such as wearing proper personal protective equipment (PPE), proper marking of danger zones, and maintaining exclusion zones, construction site inspectors must walk around sites to identify hazards, document hazards (e.g., take photos), propose solutions, and write reports. This traditional yet prevalent practice is time-consuming, labor-intensive, and does not guarantee comprehensive coverage of all safety violations due to individual judgments.

To reduce the demand for time and manpower, the construction industry has been actively employing artificial intelligence (AI) models based on deep learning, a subset of machine learning that uses multi-layered neural networks to automatically learn hierarchical patterns from data. Early implementations of deep learning focused on specialized vision-only architectures such as convolutional neural networks (CNN) for visual recognition in safety inspection tasks, such as PPE detection or pose estimation, through on-site images and videos \citep{Chen2021PPE,Xiong2021PPE,An2021MOCS}. The novel transformer architecture makes it possible for AI models to effectively understand long paragraphs of text. Derived from this advancement, VLMs are a type of deep learning model that fuse the vision and language modalities to understand both image and text inputs. The construction industry has been actively incorporating VLMs with supervised training into various safety inspection tasks, such as PPE detection \citep{Gil2024PPEImageCaption}, construction scene captioning \citep{Liu2020manifesting,Xiao2022Feasibilitystudy,Jung2024VisualSiteDiary,Kim2025LVLM}, safety rule violation detection \citep{Fang2023IEEE,Liu2022Grounding,Kim2025LVLM,Chan2025VLMontology}, and safety report generation \citep{Tsai2023Safetyclip}.

Although previous studies have demonstrated the feasibility of training VLMs with construction site images for specialized tasks, the small model size and limited training data, both in size and task type, hinder the models' ability to generalize to further downstream tasks for which they were not explicitly trained, creating a \textbf{modeling challenge} \citep{Liu2020manifesting,Fang2023IEEE,Liu2022Grounding}. Moreover, the lack of a comprehensive dataset means researchers have to take photos of individual construction sites, design their own tasks, and annotate their own data, constituting a \textbf{data challenge} \citep{Gil2024PPEImageCaption,Liu2020manifesting,Xiao2022Feasibilitystudy,Fang2023IEEE,Tsai2023Safetyclip,Chan2025VLMontology,Kim2025LVLM}. Data augmentation techniques have been adopted to mitigate data scarcity in prior works. \cite{Tsai2023Safetyclip} fine-tuned a Contrastive Language-Image Pre-training (CLIP) model for safety violation report generation and addressed class imbalance of violation data by resampling minority classes to ensure balanced fine-tuning. Similarly, \cite{Liu2020manifesting} augmented their dataset by generating five synonymous descriptions for each construction site image, thereby increasing the diversity of textual captions. However, these methods often fail to provide more training opportunities about domain-specific challenges in construction safety tasks, such as new spatial relationships in cluttered environments. Some researchers have used VLMs to generate training data \citep{Chan2025VLMontology,Kim2025LVLM,Jung2024VisualSiteDiary}. However, such data is often either small in scale, limited in task diversity, or lacks human verification. This insular landscape hinders the development of new models and their application in construction safety tasks. In this work, we attempt to tackle both the modeling and data challenges.

With recent breakthroughs of large pre-trained VLMs (e.g., GPT family and LLaVA) in a variety of everyday language and vision tasks, it is natural to ask: \textit{Can pre-trained VLMs be directly deployed to improve construction safety through automated inspections?} The rationale behind this query is the flexibility provided by large pre-trained VLMs such as Gemini. For instance, in everyday object detection tasks, there is no need to restructure or retrain the model to detect a new object category—a simple prompt adjustment can address the issue due to the powerful generalization capabilities of large models. Moreover, the in-context learning capacity means that models can quickly output in the format desired given few examples, also called a few-shot setting. Additionally, large pre-trained VLMs are able to generate language output, which can be later parsed into other formats (e.g., bounding boxes). Therefore, these models have the potential to be applied to a broader range of construction tasks, such as generating safety violation reports, as compared to other specialized models. 

To make better use of these models, it is crucial to understand the perception and decision-making capacities of large pre-trained VLMs. With this crucial task in mind, we designed three experiments to benchmark large VLMs' capabilities for construction safety inspection. These experiments include: (1) describing images in natural language to demonstrate image understanding, (2) safety rule violation VQA, i.e., giving a model an image-text pair and query the model if there is any safety rule violations, and (3) construction element visual grounding, i.e., locating construction elements in an image based on a natural language query. A high-level schematic diagram showing the nature of three tasks are presented in Figure \ref{fig:summary workflow}. To the best of our knowledge, there has been no research focused on evaluating large pre-trained VLMs in the context of understanding construction site images. Specifically, it remains to be determined whether these models can be effectively utilized for downstream tasks without further supervised training. 

To implement these tasks, we first curated a dataset we named \textit{ConstructionSite 10k}, which consists of 10,013 construction site images of varying quality and topics, accompanied by multiple types of human labeled annotations. These annotations include image captions; safety rule violation, reasoning, and grounding; and construction site element visual grounding. \textit{ConstructionSite 10k} addresses the lack of image captioning datasets with comprehensive description of site images. The dataset also introduces downstream tasks such as VQA and visual grounding tailored specifically for construction applications. \textit{ConstructionSite 10k} can serve as a pre-training dataset for VLMs to understand construction sites and construction related tasks or as a fine-tuning dataset for VQA or visual grounding purposes. We tested several popular large pre-trained VLMs, including the state-of-the-art (SOTA) Gemini series and the open-source LLaVA series, in both zero-shot and few-shot settings to determine whether these high-performing pre-trained models already possess the ability to generalize their knowledge and apply it to construction safety inspection.

Testing results show that the latest large proprietary models such as GPT and Gemini can understand construction related objects in images, including their attributes, interactions, and background, underscored by thoroughly generated image captions, and setting up the SOTA in the image captioning task. Smaller open-source VLMs do not perform as well as proprietary models. In the safety violation VQA section, VLMs score high in recall but low in precision, as compared to ground truth labeled by humans. Nonetheless, none of the VLMs achieved desirable results in any of the visual grounding tasks. Performance degrades to an Intersection over Union (IoU) below 20\% for abstract regions (e.g., safety violation regions), objects with irregular shapes, and specific types of objects within a category (e.g., workers \textit{with white hard hat} rather than any workers). While VLMs, including the SOTA Gemini series, CURRENTLY still require improvements to be fully competent in construction inspection tasks, \textit{ConstructionSite 10k} and the evaluation framework proposed in this paper lay the groundwork for further training and deployment of large pre-trained VLMs in construction-related applications.

We summarize our contribution as the following:

\begin{enumerate}
		\item We propose \textit{ConstructionSite 10k}, which  contains 10,013 images taken at different construction sites and annotations containing image caption, safety violations VQA, and object visual grounding. 
		\item We propose a framework using image captioning, three-stage safety rule violation VQA, and construction element visual grounding to probe VLMs' understanding of construction site images and their capacity to follow the prompt and perform construction safety inspection tasks.
		\item We perform comprehensive experiments to evaluate state-of-the-art, off-the-shelf large pre-trained generative VLMs on their ability to generalize their knowledge by performing construction-related tasks in a zero-shot and five-shot settings.
\end{enumerate}

\section{Background}
\label{background}

\subsection*{Vision Language Models}

VLMs have become prominent due to their remarkable capabilities in understanding both images and human languages. Among various efforts to connect the visual and linguistic worlds, \cite{Radford2021learning} stand out as pioneers with their creation of CLIP. The CLIP model, which utilizes both an image encoder and a text encoder, was pre-trained on image-text pairs to understand images, language, and their relationships. Building on the pre-trained image encoder from CLIP and the open-source LLaMA language model, \cite{Liu2023llava} and \cite{Liu2024improvedllava} fine-tuned their LLaVA model using instruction-following data. This visual instruction tuning approach transforms a VLM from a next-word predictor to a visual assistant capable of following instructions and responding to requests. Their understanding of images and text, along with their ability to follow prompts, makes them promising tools for performing construction site inspections. 

Moreover, VLMs' in-context learning ability enables them to be quickly applied to different construction inspection tasks in zero-shot (i.e., instruction only) and few-shot (i.e., instruction and some examples in human language) settings without additional supervised training or architectural change. The models quickly understand the task with merely a description of the task. GPT-2, notable for its strong unsupervised task learning capabilities, showcased SOTA performance across various benchmarks despite being under-fitted to the WebText dataset \citep{Radford2019GPT2}. Building on this, GPT-3 \citep{Brown2020GPT3} further illustrates that large pre-trained models can achieve performance improvements in few-shot settings. The success of zero-shot and few-shot inferences underscores the superior generalization capacity of large pre-trained models that excel in domains they are not trained for without the need for human-labeled datasets or modifications to their architecture.

To improve the reasoning capability of large models, Chain-of-thought (CoT) prompting was proposed by \cite{Wei2023CoT}. CoT introduces intermediate reasoning steps when querying large models with complex questions. This approach encourages models to break down a complex task into smaller, more manageable parts and think through each segment in detail. 

In this paper, we investigate the core question of whether current VLMs are capable of understanding construction sites and conduct construction safety inspection through in-context learning. We achieve this by evaluating the capabilities of both larger and smaller VLMs through image captioning tasks and safety rule violation tasks across zero-shot, few-shot settings, and CoT scenarios. Our objective is to verify whether the idea of leveraging in-context learning can inform strategies for future training and inference of large pre-trained models. This exploration not only tests the robustness of these models but also explores avenues for optimizing their application in real-world scenarios.

\subsection*{Applications of VLMs in construction}

A summary of recent works of applying VLMs in the construction industry, as well as our own work, is presented Table \ref{tab:summary of VLM in construction}. The ability of VLMs to understand both images and human language facilitates their deployment in inspection tasks within construction sites. For instance, in enhancing PPE compliance detection, \cite{Gil2024PPEImageCaption} leveraged the image captioning capability of the CLIP model. This involves describing cropped regions of construction site images containing human bodies with a image captioning model derived from CLIP. The resulting text embeddings were then used to calculate cosine similarity with prompts indicating compliance or violation to determine rule adherence. Remarkably, their proposed zero-shot method outperformed traditional YOLO methods in this context. 

To broaden the application of VLMs beyond PPE detection, \cite{Fang2023IEEE} proposed a VLM architecture comprising a Faster R-CNN as the image encoder and a GRU network as the text encoder. Their model was trained to detect eight different types of safety rule violations in construction site images. Expanding on this work, \cite{Liu2022Grounding} further developed the research by incorporating a model capable of grounding these violations with bounding boxes. They employed DarkNet as the image encoder and BERT as the text encoder.

Generative VLMs can be used to describe the construction scene and aid in construction management tasks. Both \cite{Liu2020manifesting} and \cite{Xiao2022Feasibilitystudy} developed generative VLMs for this purpose. \cite{Liu2020manifesting} trained a model consisting of a CNN image encoder and an LSTM text decoder. This model generated structured descriptions by selecting words from a pre-defined descriptive sentence bank. On the other hand, \cite{Xiao2022Feasibilitystudy} employed attention mechanisms in their model, focusing specifically on describing construction vehicles. Additionally, \cite{Tsai2023Safetyclip} aimed to use VLMs to automate the generation of safety violation reports for superior efficiency and consistency. To achieve this, they fine-tuned a variant of the CLIP model with two separate text encoders: one encoding ``violation type" and the other ``caption type". These encoded choices were used to calculate cosine similarity with the encoded image to determine the correct attributes. Subsequently, the encoded attributes, image, and image caption embeddings were fed into a GPT-2 model to generate safety violation reports with a standardized format.

To further advance VLM applications in construction, several studies have explored domain-specific fine-tuning. Both \cite{Jung2024VisualSiteDiary} and \cite{Kim2025LVLM} adapted pre-trained VLMs using construction site image caption datasets. \cite{Jung2024VisualSiteDiary} introduced the first vision transformer–based image captioning model for construction, fine-tuning an mPLUG–BERT architecture \citep{Li2022mplug,Devlin2019BERT} to generate domain-specific captions that support downstream tasks such as report generation and image retrieval. \cite{Kim2025LVLM} fine-tuned the LLaVA-1.5 image encoder to emphasize construction activities, enabling the model to produce safety-aware captions that describe the scene, assess safety compliance, and provide justifications. Similarly, \cite{Chan2025VLMontology} fine-tuned CogAgent \citep{Hong2024CogAgent} using construction object labels and safety ontologies, transforming it into a video-based analyzer for detecting working-at-height violations.

In this paper, we aim to overcome the limitations of task-specific models and supervised training by exploring the capabilities of large pre-trained generative VLMs in the context of construction site images. Our approach focuses on assessing these models' ability to comprehensively understand construction site images and accurately express this understanding through unconstrained image captions. Additionally, we leverage their capacity for in-context learning to tackle tasks such as VQA and visual grounding.

\begin{table*}[htb!]
	\caption{Recent works of applying VLM for construction site inspection.}
    \scriptsize
	\label{tab:summary of VLM in construction}
	\begin{tabular}{p{0.2\textwidth}p{0.2\textwidth}p{0.2\textwidth}p{0.3\textwidth}}
		\hline
		\textbf{Reference} & \textbf{Task} & \textbf{Training data} & \textbf{Dataset open source or not} \\ \hline
		\cite{Gil2024PPEImageCaption} & Image captioning, PPE detection & None (zero-shot) & Yes, test set is available upon request \\
		\cite{Fang2023IEEE} & Safety rule violation detection & 700 images & No \\
		\cite{Liu2022Grounding} & Safety rule violation detection and grounding & MSCOCO object detection dataset as pre-training data, 860 images with augmentation as fine-tuning data & Yes, fine-tuning data is available upon request \\
		\cite{Liu2020manifesting} & Image captioning & 7,400 images and 36,900 captions & Yes, on GitHub \\
		\cite{Xiao2022Feasibilitystudy} & Image captioning & 4,000 images and 8,000 captions & Yes, data is available upon request  \\
		\cite{Jung2024VisualSiteDiary} & Image captioning, Safety report generation & 6,400 images and 13,000 captions & Yes, on GitHub \\
		\cite{Kim2025LVLM} & Image captioning, safety rule violation detection & 2,000 images, 2,000 cations, and 2,000 safety rule VQA & Yes, data is available upon request \\
		\cite{Chan2025VLMontology} & Video-based safety rule violation detection & 900 images, 900 captions, and 900 safety rule VQA & Yes, data is available upon request \\
		\cite{Tsai2023Safetyclip} & Safety report generation & 806 image-text pairs with augmentation as fine-tuning data & No \\ \hline
		\textbf{\textit{\textbf{Ours}}} & image captioning, multiple choice, question answering, visual grounding & None (zero-shot and few-shot) & Yes, both training set and test set is available online \\ \hline
	\end{tabular}
\end{table*}

\subsection*{Multi-modal Datasets}

There have been extensive efforts to compile multi-modal datasets aimed at training VLMs to effectively integrate vision and language modalities. Among the popular datasets are Flickr \citep{Hodosh2013Flickr8k}, MSCOCO \citep{Chen2015MSCOCOcaptions}, and Abstract Scene \citep{Zitnick2013AbstractScene} datasets. These datasets contain large volumes of images, each accompanied by multiple captions, and have been widely utilized as pre-training datasets for VLMs. RefCOCO \citep{Kazemzadeh2014referitgame}, pioneering in visual grounding, plays a crucial role in this domain by enabling the detection of objects through language prompts provided to VLMs. Visual grounding, distinct from traditional object detection, allows for additional constraints (e.g., color, location) to be imposed on the detection task. Furthermore, VQA datasets such as VizWiz \citep{Gurari2018VizWiz}, OKVQA \citep{Marino2019okvqa}, and ScienceQA \citep{Lu2022ScienceQA} significantly expand the utility of VLMs beyond image description to answering questions related to images, which broadens the scope of applications for these models. 

Datasets containing images with alt-text style captions, tend to be straightforward. These images often have excellent illumination, fewer objects, and simple layouts. In contrast, images captured at construction sites present significant challenges due to poor weather conditions, varying illumination, and the complex perspectives from UAVs or tower cranes, often at a distance. Moreover, these images contain rich content with multiple objects from diverse classes and various interactions, posing considerable confusion for machine learning models. The differences between these types of datasets render VLMs trained on common object-centric datasets less effective in construction-related tasks. 

While there are publicly available datasets containing construction site images such as SODA \citep{Duan2022SODA}, MOCS \citep{An2021MOCS}, and ACID \citep{Xiao2021ACID}, their primary focus is typically on object detection tasks. The CASIC\footnote{Unofficial abbreviation for “Construction Activity Scene Image Caption”.} dataset \citep{Liu2020manifesting} aims to address the gap by providing captioned construction site images. However, it covers only a limited range of activities with an average of merely 12 words per caption. The VSD dataset \citep{Jung2024VisualSiteDiary} tried to create a image captioning dataset with natural language but relied on uncorrected machine-generated captions. The captions also only provide summary of the entire scene without diving into details. Table \ref{tab:construction datasets comparison} compares our dataset with the construction-related computer vision datasets aforementioned to better visualize the similarities and differences between the datasets.

This deficiency underscores the need for specialized datasets that comprehensively capture the complexities of construction site environments and capture as many semantic features as possible from images. Such datasets would enable VLMs to better understand the context surrounding the main scene and establish more nuanced relationships between objects. Moreover, datasets focusing on downstream tasks like safety rule violation VQA and visual grounding in construction site inspection are also lacking. 

To address these gaps, this paper proposes the creation of an image captioning dataset aimed at enhancing VLM training. Our dataset also introduces VQA and visual grounding tasks specific to construction site inspection. This initiative seeks to foster advancements in VLMs' capabilities within the challenging contexts of construction environments.

\begin{table}[h!]
\centering
\caption{Comparisons between our dataset with popular construction-related computer vision datasets.}
\label{tab:construction datasets comparison}
\scriptsize
\resizebox{\textwidth}{!}{
\begin{tabular}{p{0.15\textwidth}p{0.07\textwidth}p{0.17\textwidth}p{0.08\textwidth}p{0.15\textwidth}p{0.12\textwidth}p{0.3\textwidth}}
\hline
\textbf{Dataset} & \textbf{No. of Images} & \textbf{Type of Annotations} & \textbf{Avg. Caption Length} & \textbf{No. of Question-Answer Pairs per Image} & \textbf{Object Category for Detection} & \textbf{Features} \\ \hline
\makecell[l]{\textbf{SODA}\\ \citep{Duan2022SODA}} & 20,000 & bounding box & - & - & 15 & \parbox[l]{0.3\textwidth}{1. object detection at construction sites} \\ 
\makecell[l]{\textbf{MOCS}\\ \citep{An2021MOCS}} & 42,000 & \makecell[l]{bounding box, \\ segmentation mask} & - & - & 13 & \parbox[l]{0.3\textwidth}{1. object detection and instance segmentation \\ at construction sites \\ 2. collected from a variety of construction projects} \\ 
\makecell[l]{\textbf{ACID}\\ \\ \citep{Xiao2021ACID}} & 10,000 & bounding box & - & - & 10 & \parbox[l]{0.3\textwidth}{1. focusing on construction machines} \\ 
\makecell[l]{\textbf{CASIC}\\ \citep{Liu2020manifesting}} & 7,000 & caption & 11.8 & - & - & \parbox[l]{0.3\textwidth}{1. each image focuses on only one construction activity \\ 2. each image is described with five similar short sentences} \\ \\
\makecell[l]{\textbf{VSD}\\ \citep{Jung2024VisualSiteDiary}} & 7,800 & caption & 10.3 & - & - & \parbox[l]{0.3\textwidth}{1. training and validation data includes diverse actions and jobsite environments to prevent overfitting \\ 2. some captions are generated using pre-trained VLM and validated against annotated object existence} \\ \\ \hline
\textbf{ConstructionSite 10k} & 10,000 & \makecell[l]{caption,\\ bounding box,\\ multiple choices, \\ one or two sentences \\ for safety rule VQA,\\ bounding box} & 53.0 & 3 & 3 & \parbox[l]{0.3\textwidth}{1. detailed image captions for construction site images \\ 2. captions contain predicates indicating relationships \\ between instances \\ 3. relative and absolute positional information contained \\ 4. three-stage VQA \\ 5. visual grounding under constraints} \\ \hline
\end{tabular}
}
\end{table}

\subsection*{Research objective}

We hypothesize that if VLMs can grasp every detail and relationship within a construction site image, they can be applied to various tasks simply by adjusting the input prompt. To test our hypothesis, we have curated a comprehensive construction image caption dataset, \textit{ConstructionSite 10k}, and developed a framework to evaluate the VLMs' capacity to perform multiple construction site-related tasks under zero-shot and few-shot settings. The proposed framework includes three tasks: image captioning, safety rule violation VQA, and visual grounding tasks. It also includes metrics used to evaluate models' performance on each task, as well as prompts and hyperparameters used.

\section{ConstructionSite 10k}
\label{Dataset}

This section presents a comprehensive analysis and visualization of the \textit{ConstructionSite 10k} dataset. The dataset is publicly available at: https://huggingface.co/datasets/LouisChen15/ConstructionSite

\subsection*{Images}

The construction site images utilized in this dataset were collected and made publicly available by \cite{An2021MOCS}. We selected a total of 10,013 images that are clear, have acceptable illumination, and contain distinguishable construction elements that are not occluded by mosaics. 

Images captured across various construction sites, times, and with different equipment will exhibit distinct characteristics. Building on the groundwork laid by \cite{An2021MOCS}, we assign four key features to each image to reflect its condition: Camera distance, Illumination, Plan view, and Quality of information. These labels are intended to aid users in selecting images suitable for specific purposes. 

Within this framework, camera distance refers to the proximity of the camera to the majority of construction elements in the image. Illumination describes the lighting conditions present in the image. An image is classified as a plan view if the angle of view exceeds approximately 45 degrees from the ground. Images with sparse information typically depict isolated instances or small groups of construction elements such as humans, equipment, and material stockpiles, each clearly distinguishable without significant occlusion. In contrast, images labeled as rich information portray a diverse array of construction elements with complex interactions and occlusions between them. 

\begin{figure*}[ht!]
	\centerline{\includegraphics[width=0.8\textwidth, height=0.6\textheight]{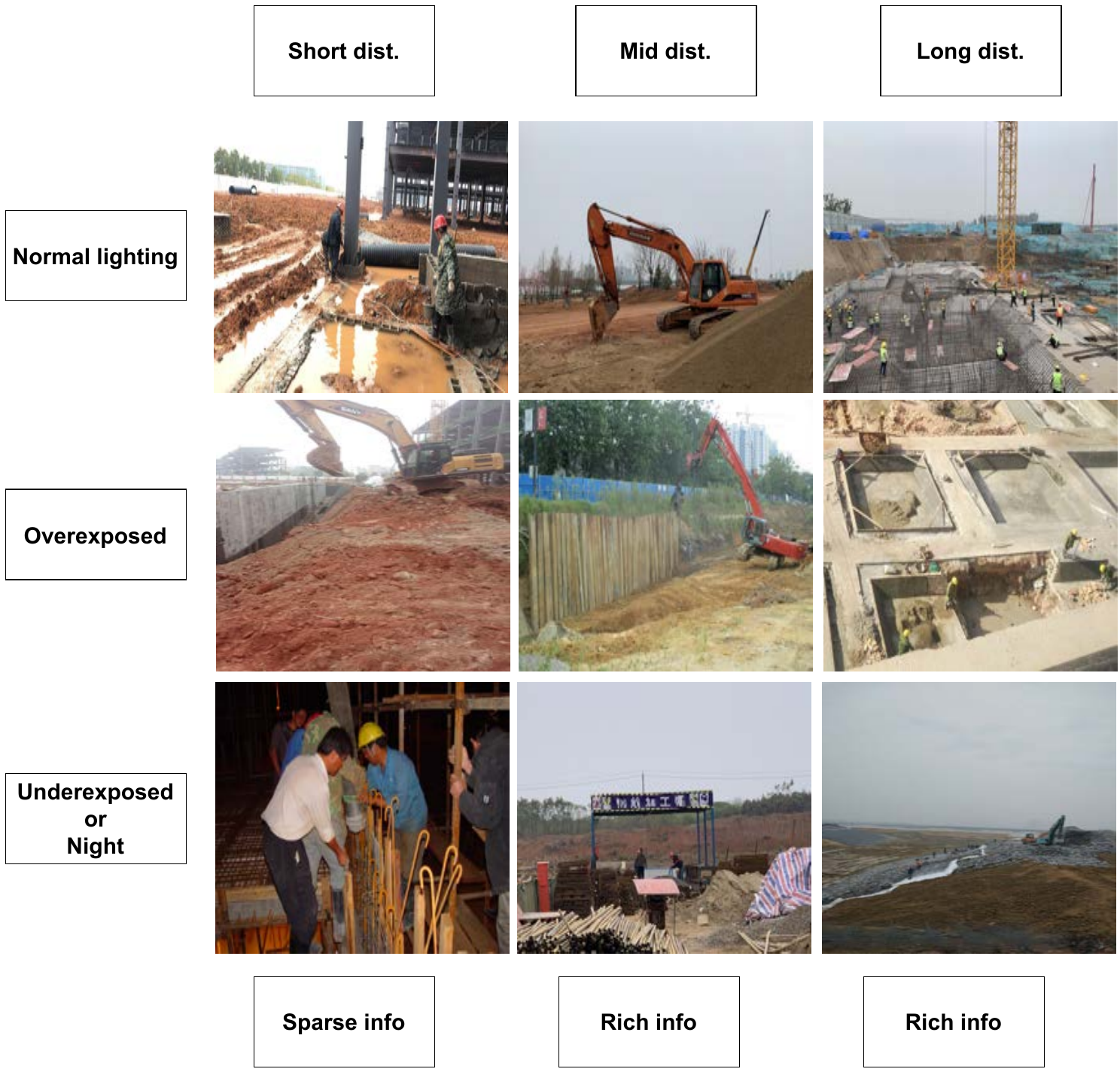}}
	\caption{The figure presents construction site images from the dataset, each annotated with three labels arranged in a grid structure. For example, the bottom-left image is labeled as ``sparse info'', ``night'', and ``short distance''. The images are uniformly scaled for demonstration purposes, resulting in slightly altered appearances. These images and labels aim to showcase representative samples and do not depict the actual distribution of the dataset.}
	\label{fig:image and annotation demo}
\end{figure*}

\begin{figure}[htb!]
	\includegraphics[width=0.8\textwidth]{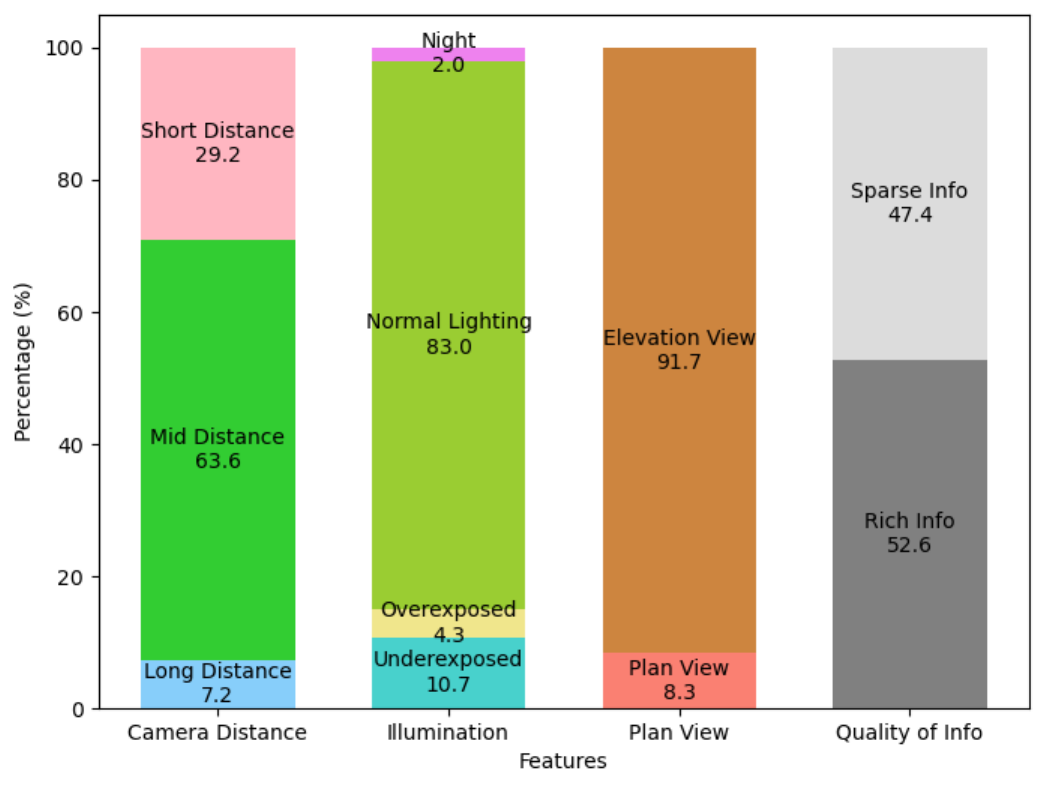}
	\caption{Distribution of images in the dataset across four features. These statistics unveil the diversity of construction site imagery.}
	\label{fig:dataset features bar chart}
\end{figure}

The assignment of these labels follows predefined criteria but also incorporates annotator judgment. Figure \ref{fig:image and annotation demo} provides visual examples showcasing images with different feature labels. Figure \ref{fig:dataset features bar chart} illustrates the distribution of these features across the dataset. These features offer useful references for researchers who want to divide the dataset to create challenging subsets for model training. This study represents one of the earliest attempts to systematically describe image conditions through assigned labels.

\subsection*{Image annotations and corresponding tasks}

\begin{figure}[htb!]
    \centering
    \includegraphics[width=0.75\textwidth]{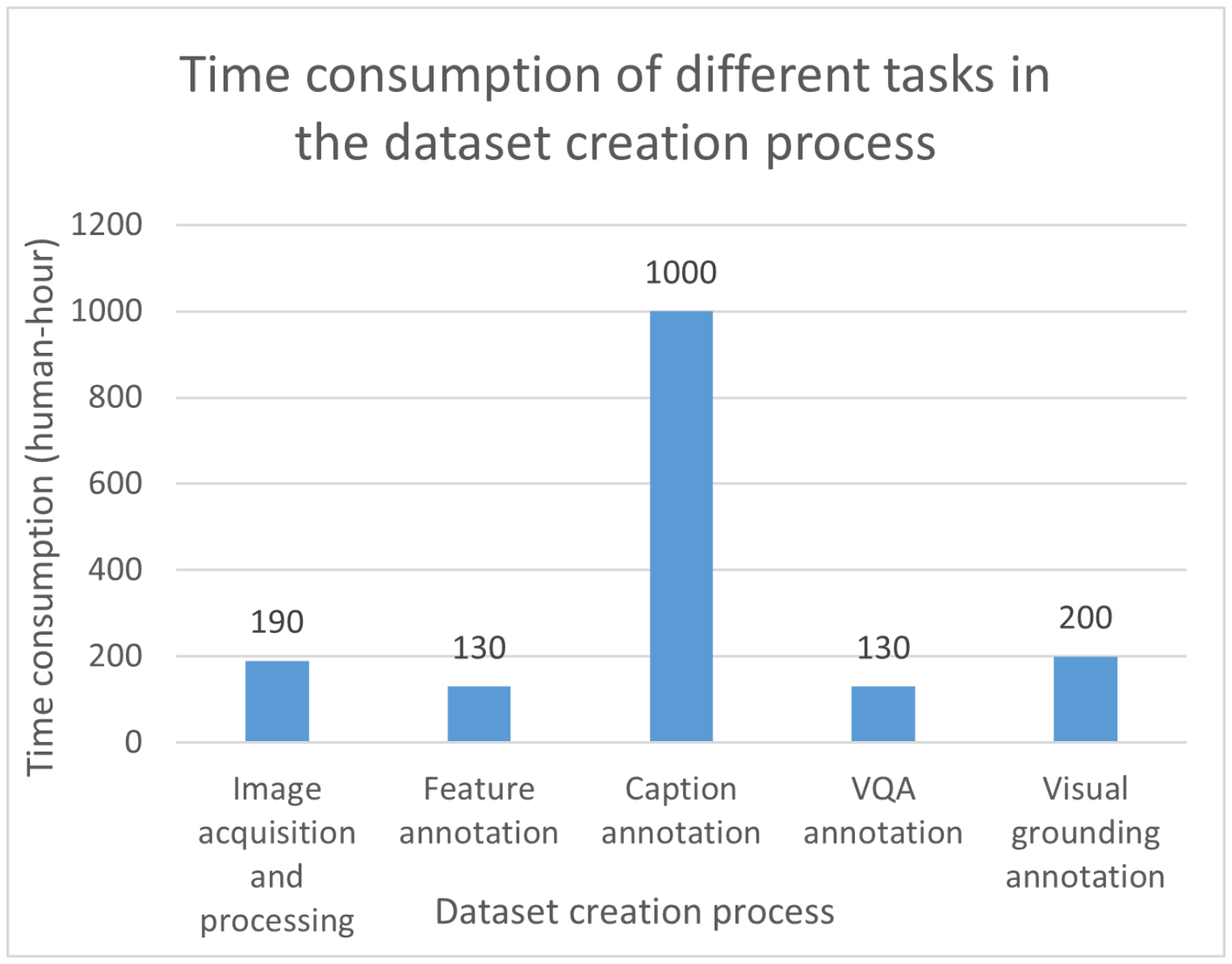}
    \caption{The time consumption breakdown to prepare the dataset. The time shown in the figure approximates and includes the time to create annotation software.}
    \label{fig:time consumption}
\end{figure}

The images were manually annotated by the authors, with various types of annotations corresponding to different tasks. All annotations were reviewed and validated by two domain experts- construction engineers with experience in both construction site safety inspections and academic research. The following subsections are categorized by tasks used to evaluate VLMs. The time used to create the dataset and prepare the annotations are presented in Figure \ref{fig:time consumption}. 

\begin{figure}[htb!]
	\centerline{\includegraphics[width=\textwidth]{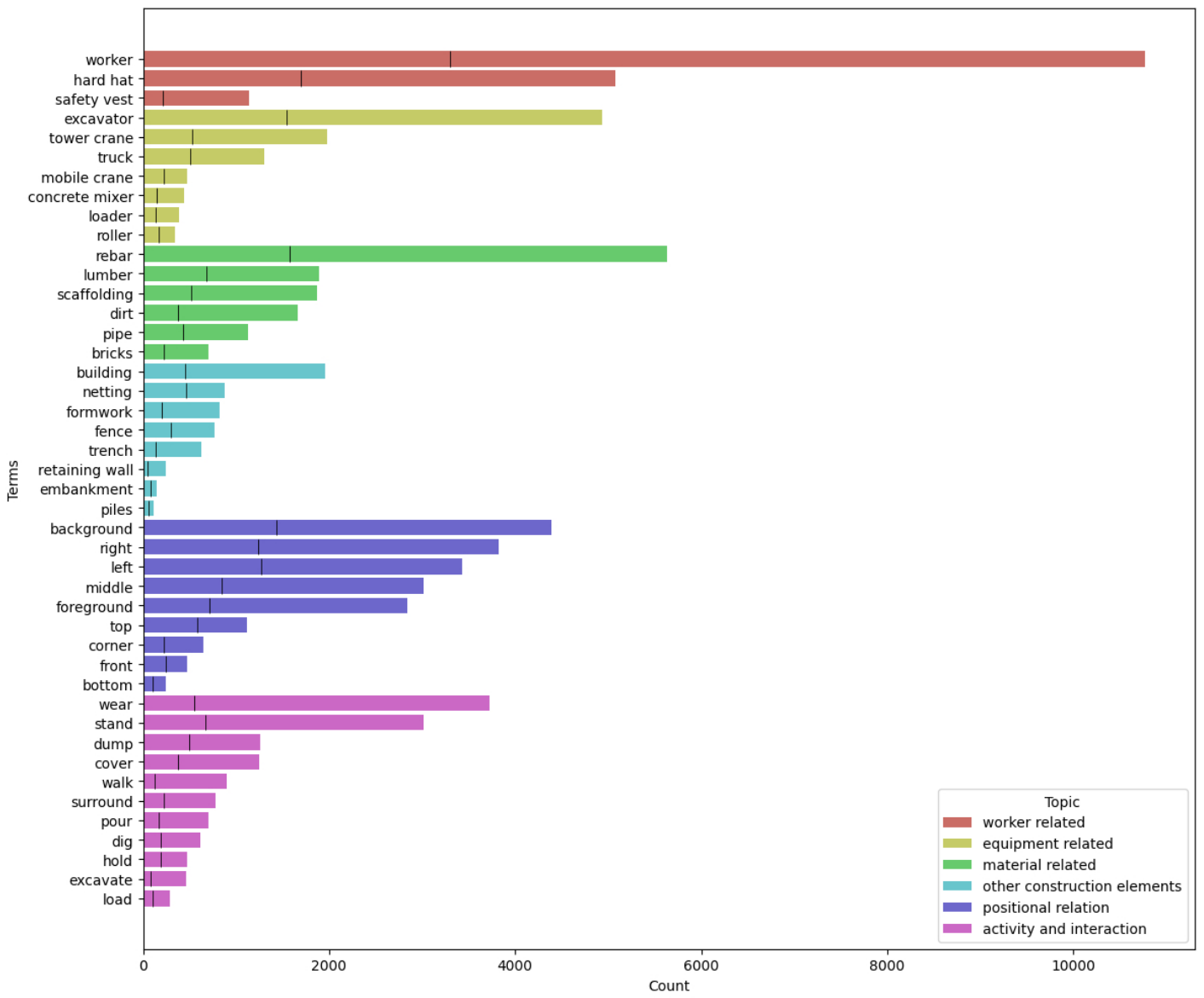}}
	\caption{Word frequency of prevalent terms across topic categories in the reference image captions. Synonymous and plural noun forms (e.g., concrete truck and concrete mixers) are consolidated under a single canonical term. Verb occurrences are aggregated at the lemma level. For each term, the bar shows the total frequency, with occurrences from the test split on the left and training split on the right, separated by a black vertical marker.}
	\label{fig:word count}
\end{figure}

The dataset was split into a training set comprising 70\% of the images and corresponding annotations, and a test set comprising the remaining 30\%. More detailed split of annotations is available in Figure \ref{fig:word count} and Table \ref{tab:annotation statistics}. The training set is intended for model training and fine-tuning, while the test set is exclusively used in this study to evaluate the performance of large pre-trained VLMs.

\textbf{Image captioning. }

\begin{figure}[htb!]
	\centerline{\includegraphics[width=\textwidth]{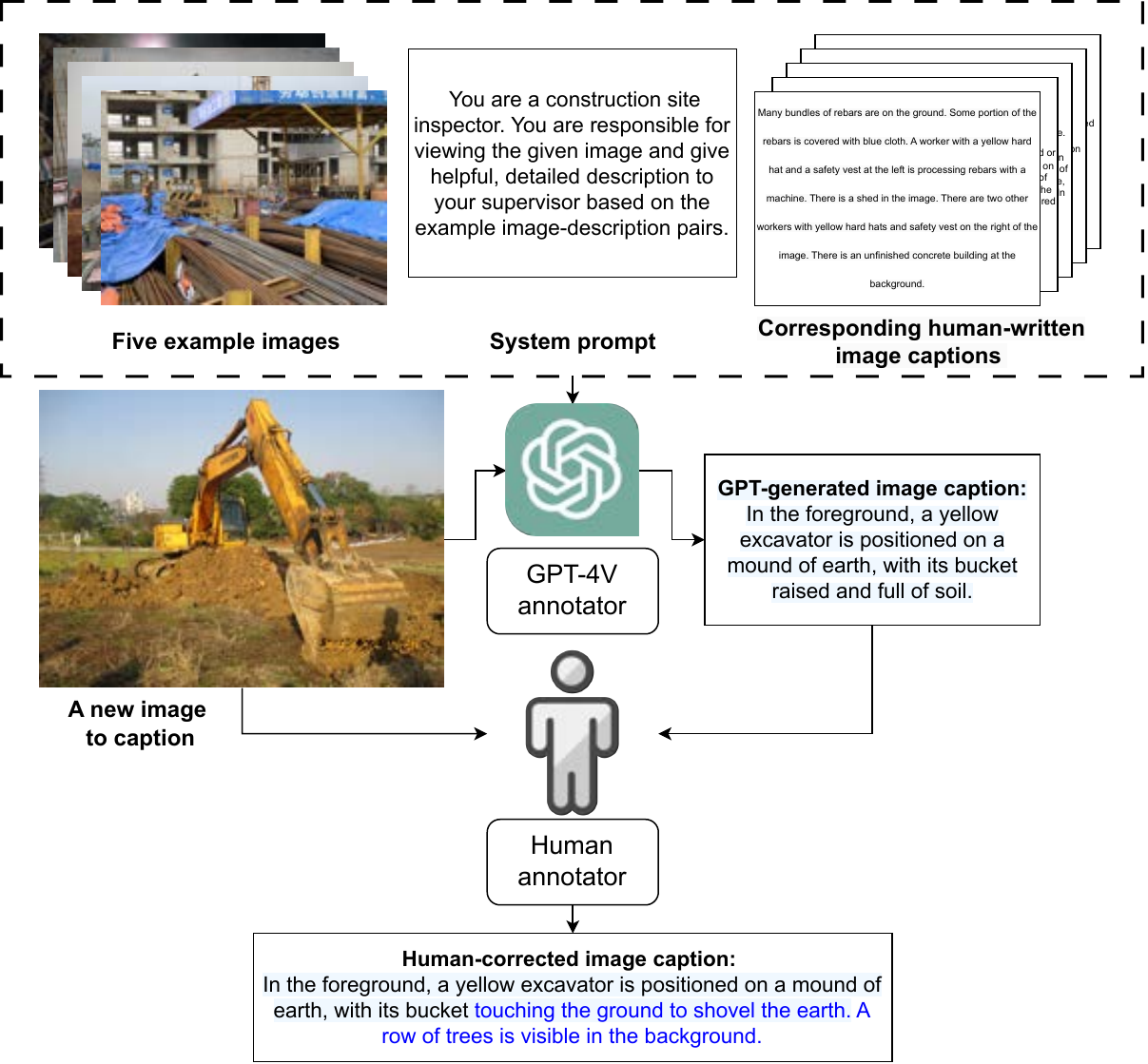}}
	\caption{The overall workflow of GPT-assisted image caption annotation.}
	\label{fig:image captioning annotation example}
\end{figure}

The annotations include detailed and objective descriptions in English with a focus on construction elements such as workers, construction equipment, and materials. Attributes of objects such as color, location, number, and activities are included in the image captions. We first manually annotated around 2,000 images. To increase the annotation speed, we utilized GPT-4V in a five-shot setting to create the structure of the captions for the remaining images. Human annotators then adapted the machine-generated descriptions to ensure their fidelity and adequacy. The GPT-aided approach result in an approximately 20\% reduction of annotation time for each caption. A workflow of the GPT-assisted image caption annotation is shown in Fig. \ref{fig:image captioning annotation example}. The process involves three key steps: (1) inputting task-specific prompts, few-shot examples, and a query image into the GPT-4V annotator; (2) obtaining an initial AI-generated caption; and (3) refining this caption through human annotation to ensure accuracy and relevance. This hybrid approach leverages the efficiency of GPT-4V while prioritizing human oversight to produce high-quality, domain-specific annotations.

Unlike traditional image captioning datasets, our annotations include not only the most visible objects in the foreground, but also construction elements scattered throughout the image, including those in the corners. Background details are also included as an attempt to capture every possible detail. A summary of frequently used words in the image captions is presented in Figure \ref{fig:word count}, showcasing the diverse types of objects and activities described in the dataset. 

\textbf{Safety rule violation VQA and Object visual grounding. }
\label{dataset: safety rule violation VQA}

To further evaluate VLMs' understanding of construction sites and facilitate the training of VLMs for downstream applications, we introduce safety rule violation VQA annotations. This task plays a pivotal part in training and evaluating proficiency in processing an overloading amount of visual and textual information while still strictly following the prompt and generating human-readable responses. Prior works \citep{Fang2023IEEE,Liu2022Grounding} have identified about ten common safety concerns at Chinese construction sites. However, differences between the datasets and the AI models used in their works and our paper necessitated adjustments of safety rules to ensure relevance. We adopted three safety rules from their works: ``edge protection", ``usage of safety harnesses," and ``workers within excavator operating zones", which align more with our dataset’s context. To address regional specificity, we incorporated guidelines from \cite{WorksafeBC}, which emphasize rigorous PPE requirements (e.g., full-body coverage for shoulders, legs, and toes). The four rules span PPE adherence, safety harness usage, edge protection, and human-machine interaction hazards. These categories collectively address the majority of \textbf{visually identifiable} safety risks in our dataset, balancing breadth and annotation feasibility. The rules are presented in Table \ref{tab:four safety rules}. Each image annotation includes all observed violations of the safety rules, accompanied by one to two sentences explaining who and where the rule was violated, along with bounding boxes that ground the violators. This visual grounding task differs from traditional visual grounding since the object to ground is not given in the prompt, but needs to be determined by the VLMs based on the image, safety rules, and their prior answers. 

\begin{table*}[htb!]
	\centering
	\caption{Four safety rules used in the dataset.}
	\label{tab:four safety rules}
	\begin{tabular}{p{0.15\linewidth}p{0.75\linewidth}}
		\hline
		Safety Rule ID & Description of the rule \\ \hline
		\vspace{0.5em} & \vspace{0.5em} \\
		Rule 1 & Use of basic PPE when on foot at construction sites. (hard hats, properly worn clothes covering shoulders and legs, shoes that can cover toes, high-visibility retroreflective vests at night, face shield or safety glasses when cutting, welding, grinding, or drilling). \\
		\vspace{0.5em} & \vspace{0.5em} \\
		Rule 2 & Use of safety harness when working from a height of three meters and the edges are without any edge protection. \\
		\vspace{0.5em} & \vspace{0.5em} \\
		Rule 3 & Adoption of edge protection or edge warning including guardrails, fences, for underground projects three meters in depth with steep retaining wall and for human to stand. \\
		\vspace{0.5em} & \vspace{0.5em} \\
		Rule 4 & Appearance of worker in the blind spots of the operator and within the operation radius of excavators in operation, or excavators with operators inside. \\ 
		\vspace{0.5em} & \vspace{0.5em} \\ \hline
	\end{tabular}
\end{table*}

\textbf{Visual grounding. }
\label{dataset: visual grounding}

Given the dominant text-based annotations, we consider VLMs to be only employed at construction sites if they are competent enough to ground the required object precisely. Thus, we include the construction element visual grounding section in this dataset. The visual grounding annotations include the bounding boxes of three types of construction elements: excavators, rebars, and workers with white hard hats. These elements were chosen because they are commonly seen at construction sites and represent the three crucial construction components. These three objects present increasing levels of difficulty for VLMs. Excavators are regular in shape and large in size, while workers with white hard hats introduce a specific constraint within the broader category of workers. The statistics of these annotations are shown in Table \ref{tab:annotation statistics}.

\begin{table}[htb!]
	\centering
    \scriptsize
	\caption{Statistics of the annotations. For image captioning, the table reports the number of images, sentences, and words. For safety rule violation VQA and object visual grounding, it shows the number of ground truth annotations (occurrences).}
	\label{tab:annotation statistics}
		\begin{tabularx}{0.5\textwidth}{lll}
			\hline
			Image Captioning & Test set & Training set \\ \hline
			Image count             & 3,004 & 7,009                \\
			Sentence count             & 10,754 & 21,189                \\
			Word count  & 145,674  & 385,504                \\ \hline
			Safety Rule Violation VQA &  &  \\ \hline
			PPE             & 323 & 677                \\
			Safety harness  & 25  & 59                \\
			Edge protection & 63  & 109                \\
			Blind spot      & 24  &  46               \\ \hline
			Object Detection &  &    \\ \hline
			Excavator       & 1,080 & 2,415                \\
			Rebars		    & 327 & 846                 \\
			Workers with white hard hats   & 314 & 680                \\ \hline
		\end{tabularx}
\end{table}

\section{Model selection}
We selected six large pre-trained VLMs for testing, grouped into two categories: (a) larger proprietary models that represent state-of-the-art performance, and (b) smaller open-source models that are freely available, flexible, and locally deployable. The LLaVA series, MiniGPT-4, and GPT-4 series were considered state-of-the-art in 2024, while Gemini 2.5 and Qwen 2.5 represent the state-of-the-art in 2025.

\textbf{LLaVA } \citep{Liu2024improvedllava} The LLaVA variants employed in this paper are LLaVA v1.5 13B and LLaVA v1.5 7B. LLaVA uses Vicuna v1.5 \citep{Chiang2023vicuna} as the language encoder; for the vision encoder, LLaVA uses CLIP \citep{Radford2021learning}. The LLaVA v1.6 34B variant is used for the safety rule violation VQA test. It is accessed via the Gradio server API, where we cannot modify the system prompt.\footnote{The Gradio server is accessed through: https://llava.hliu.cc/}

\textbf{MiniGPT-4 } \citep{Zhu2023minigpt4,Chen2023minigptv2} The MiniGPT-4 variant employed in this paper is MiniGPT-4 v2. The model uses an EVA \citep{Fang2022eva} as the vision encoder while the language encoder is Llama2-chat-7B. The vision encoder EVA can be considered as a scaled-up and more powerful version of CLIP. 

\textbf{Qwen-2.5 } \citep{Bai2025Qwen2.5vl} The Qwen variant used in this study is Qwen-2.5-VL (7B), selected for its state-of-the-art performance on multiple benchmarks in document understanding and VQA among open-source VLMs, as well as its strong video grounding capabilities. The model integrates a vision transformer as the vision encoder and the Qwen-2.5 language model (LLM) \citep{Yang2025Qwen2.5} as the language backbone.

\textbf{GPT-4 } \citep{Openai2024gpt4} The GPT-4 variants employed in this paper are GPT-4-1106-vision-preview (GPT4V) and GPT-4o-2024-05-13 (GPT4o). The model is chosen for its outstanding performance in image captioning and reasoning, however, we know little about its architecture, dataset, and training method.

\textbf{Gemini-2.5 } \citep{Comanici2025Gemini2.5} The Gemini-2.5 variant used in this study is Gemini-2.5-Flash, selected for its state-of-the-art performance in image captioning and reasoning. In addition, the model is reported to be further trained to convert images to structural representation and then perform image reconstructions, which may enpower it with enhanced spatial understanding capabilities.

\textbf{Grounding DINO } \citep{Liu2023groundingDINO} The Grounding DINO variant used in this paper is GroundingDINO-B with Swin-B as backbone. The model is a specialized VLM designed for open-set object detection. Although it does not generate text outputs like other VLMs, it is expected to achieve superior performance on tasks related specifically to object detection. The model serves as the upper bound for the automated construction site object visual grounding task. 

\textbf{Traditional Computer Vision (CV) Baseline (Faster R-CNN)} To provide a meaningful comparison with large pre-trained VLMs in the Object Visual Grounding task, we include a traditional computer vision baseline using Faster R-CNN \citep{Ren2015fasterrcnn} with a ResNet-50 backbone and Feature Pyramid Network (FPN). The ResNet-50 backbone was pre-trained on ImageNet, providing robust visual feature extraction. Since existing open-source models are not trained to detect challenging construction site objects (e.g., rebars) or specifically constrained objects (e.g., workers with white hard hats), we fine-tuned the model to enable detection of the categories required for our task. We trained the detection head (region proposal network and box/class predictors) from scratch on the training split of our Object Visual Grounding task. Fine-tuning consisted of 30 epochs using stochastic gradient descent (SGD) with a learning rate of $5\times10^{-4}$, momentum of 0.9, and weight decay of $1\times10^{-4}$. Data augmentation included resizing, padding, horizontal flips, color jittering, and small rotations. The resulting model serves as a quantitative baseline to evaluate the performance difference between large pre-trained VLMs and traditional computer vision models. The training procedures followed those in \cite{An2021MOCS}. Since fine-tuning a CNN-based traditional model is not the primary focus of this paper, we do not provide further details.

\section{Model evaluation}
\subsection*{Image captioning}
\label{Image captioning}

We use one system prompt and one user prompt for image captioning. The system prompt is established prior to initiating each conversation. The system prompt is: ``You are a construction site inspector. You are responsible for viewing the given image and give helpful and polite answers to your supervisor."

Each model then receives an image and a user prompt as input. The user prompt is: ``Please describe the image. Your description should include information about people, construction equipment, and material stockpiles. Do not make assumptions, be concise, and describe facts only. Please describe with only one paragraph."

We conduct queries with GPT4V, GPT4o, Gemini-2.5-VL and LLaVA 13B in a 5-shot setting. As demonstrated by previous work on in-context learning \cite{Min2022InContextLearning}, few-shot prompting has a limited impact on the underlying knowledge of large pre-trained models. Nevertheless, few-shot examples can improve performance by encouraging more consistent and well-structured outputs. Accordingly, in this study, few-shot examples are not used as teaching signals for task adaptation or performance enhancement. Instead, they serve as format-conditioning templates to guide the models toward consistent output structures, thereby boosting performance with more homogeneous response styles. The five few-shot examples are randomly sampled from the training set to avoid information leakage from the test set. For GPT-4 and Gemini models, which support inputting multiple images within a single conversation, we provide five images along with their corresponding human-annotated captions as example prompts before testing the model with the testing image. For LLaVA models, however, only one image can be inputted per conversation. Therefore, we input five captions without corresponding images for in-context learning, inspired by \cite{Alayrac2022flamingo}. 

\begin{figure*}[htb!]
	\centerline{\includegraphics[width=\textwidth]{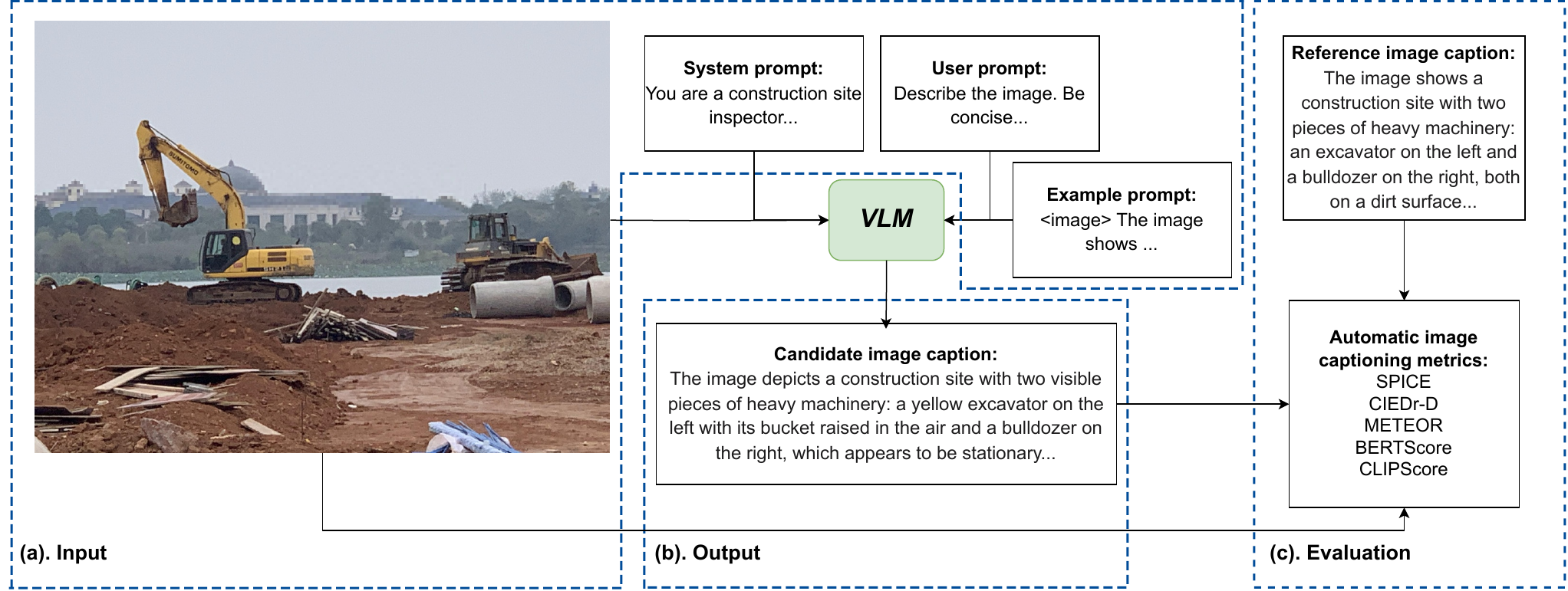}}
	\caption{Workflow of the image captioning task. The image, system and user prompts are given to the VLMs as inputs, the models generate an image caption as an output. The example prompts will be used to give examples to the VLMs in few-shot settings. The generated candidate caption, human-labeled reference caption or the image are then evaluated with automatic metrics.}
	\label{fig:image captioning workflow}
\end{figure*}

The workflow of the task and evaluation scheme is presented in Figure \ref{fig:image captioning workflow}. For GPT and Gemini models, we use a single seed due to inference cost. For open-source models, we employ five different seeds, and all reported results represent the average across these five seeds. The model's hyperparameters are set as follows: temperature=0.2, top\_p=1.0, max\_token=1024, chosen because they are either default values or closely resemble them for the employed models. 

The evaluation metrics used for the image captioning task are: SPICE, CIDEr-D, METEOR, BERTScore, and CLIPScore for their wide usage in evaluating machine-generated image captions and superior correlation with human judgment\footnotemark.
\footnotetext{Results of rule-based metrics SPICE, CIDEr-D, and METEOR are obtained using the official tools provided by MSCOCO image caption competition for comparison purposes: https://github.com/tylin/coco-caption}

\textbf{SPICE } \citep{Anderson2016SPICE} This metric parses the caption to scene graphs, which are represented by tuples, and then calculates the f1 score of candidate and reference tuples. The types of tuples employed by this metric include objects, attributes, and relationships. By converting captions into tuples, the metric focuses exclusively on the semantic meaning while ignoring the quality of text.

\textbf{CIDEr-D } \citep{Vedantam2015CIDEr} The metric calculates the similarity of the candidate caption and the consensus of one or multiple reference caption(s). It is achieved by performing a Term Frequency Inverse Document Frequency (TF-IDF) weighting for each n-gram. The equation checks the occurrence of an n-gram in a reference caption or in the candidate caption. The ``TF" term places higher weights on frequently occurring n-grams describing one image, while the ``IDF" term reduces the weights of popular words across the entire dataset since those words usually contain less unique information. 

\textbf{METEOR } \citep{Denkowski2014METEOR} This metric exhaustively maps words in the candidate caption to the reference caption according to four matchers one after another: exact token matching, stem tokens, synonyms, and then paraphrases, followed by yielding a score. A total number of matched words can be obtained.

\textbf{BERTScore } \citep{Zhang2020BERTScore} The metric makes use of an LLM to encode text to contextual embeddings and calculate the cosine similarity between candidate and reference embeddings. For this paper, we used RoBERTa-large as the embedding model. The results are re-scaled with the baseline. For a candidate caption and a reference caption, the pairwise cosine similarity is calculated between the embedding of their tokens. The maximum cosine similarity scores are then used to calculate the precision, recall, and f1 score.

\textbf{CLIPScore } \citep{Hessel2021clipscore} The metric employs CLIP \citep{Radford2021learning} to encode the images and caption texts to visual embedding and text embedding, and a cosine similarity is calculated between embedding from the two modalities followed by multiplying the result with a weight $w=2.5$. The metric enables evaluating candidate captions without references.

\subsection*{Safety rule violation VQA}
\label{benchmarking: safety rule violation VQA}

\begin{figure*}[htb!]
	\centerline{\includegraphics[width=\textwidth]{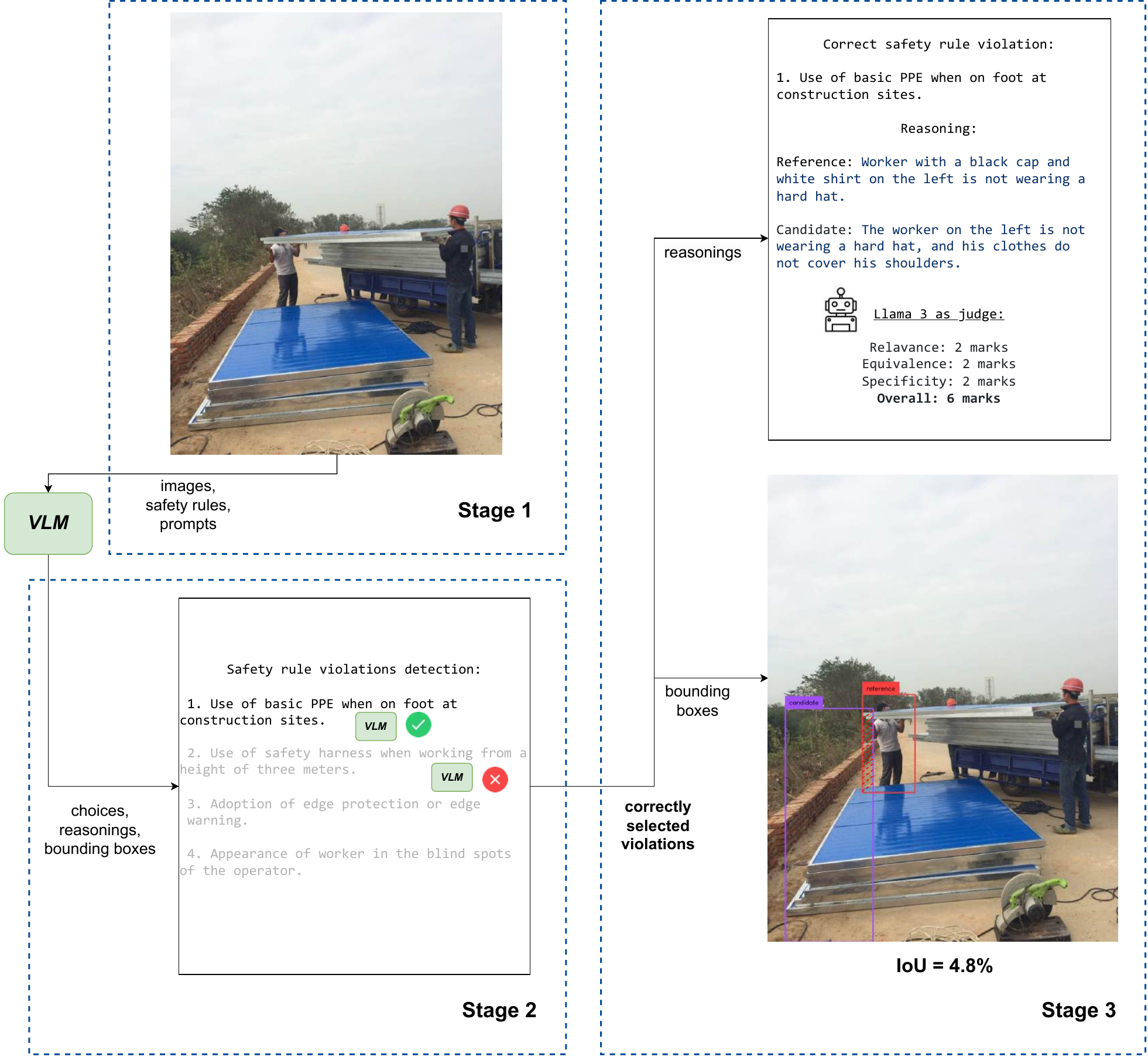}}
	\caption{A three-stage query and evaluation workflow for the safety rule VQA test. The images and prompts are given to VLM in \textbf{Stage 1}. The VLMs generate choices, reasonings, and bounding boxes as outputs. If the VLM select the correct violations in \textbf{Stage 2}, the reasonings and bounding boxes will be evaluated in \textbf{Stage 3}. The safety rules depicted in the figure are simplified for clarity, with soild font indicating rules relevant to the image and grayed-out font indicating irrelevant rules. In this example, the VLM selects \textit{Rule 1} and \textit{Rule 2}, but only \textit{Rule 1} is violated in the image. The reasoning for the violation of the correctly chosen safety rule (\textit{Rule 1}) is then provided to Llama 3 as the candidate reasoning, alongside the reference reasoning. For the visual grounding, the candidate bounding box, marking the location of the violation, is compared with the reference bounding box. This comparison yields an IoU score}
	\label{fig:safety rule violation vqa workflow}
\end{figure*}

This section consists of three tasks: multi-label classification, reasoning, and visual grounding, each detailed in Section \ref{dataset: safety rule violation VQA}. For the model evaluation, we derive a three-stage framework: the images and prompts are given to the VLMs at Stage 1; the violated rules selected by the VLMs will be checked at Stage 2; for correctly selected violations, the reasoning and bounding boxes will be evaluated at Stage 3. The workflow of the task and evaluation scheme is presented in Figure \ref{fig:safety rule violation vqa workflow}. 

Prior to beginning these tasks, we set the hyperparameters of the VLMs identical to those in the image caption task. The new system prompt is formulated as follows to ensure a structured response: ``You are a construction site safety inspector. Your responsibility is to review the provided image and provide helpful, detailed, and polite responses to your supervisor. You should only answer questions posed by the supervisor in the exact manner requested."

The model then receives a user prompt detailing the four safety rules, requesting a choice for the ID of the violated rule, an explanation for the choice, and a bounding box to ground the explanation. Additionally, we experimented with the CoT technique by querying the model independently for each of the four rules.

Due to the complexity of the proposed VQA task - which involves multi-label classification, reasoning, and bounding box generation - the correctness and consistency of answer formatting are critical for reliable evaluation. For this reason, and consistent with the rationale described in Section \ref{Image captioning}, we adopt a few-shot setting for the models of GPT4, Gemini, and LLaVA 34B as these models have sufficient context length to accommodate complex example prompts. As explained in Section \ref{Image captioning}, few-shot examples are randomly selected from the training set. In the five-shot setting, one example contains no safety rule violation, while each of the remaining examples corresponds to a different safety rule violation. This design ensures coverage of the possible input–output structures encountered during model inference. In the one-shot setting, we select an image that violates the largest number of safety rules, providing a compact yet comprehensive template for output formatting.

For multi-label classification (i.e. choosing the violated rules), the evaluation metrics are precision and recall for each rule. For the visual grounding, we use IoU. Their equations are listed in Eqn. \ref{precision}, \ref{recall}, \ref{iou}. 

\begin{equation}
	\label{precision}
	Precision =  \frac{True \; Positive}{True \; Positive + False \; Positive}
\end{equation}
\begin{equation}
	\label{recall}
	Recall =  \frac{True \; Positive}{True \; Positive + False \; Negative}
\end{equation}
\begin{equation}
	\label{iou}
	IoU =  \frac{Area \; of \; overlap}{Area \; of \; union}
\end{equation}

For evaluating the quality of reasoning, we used three criteria:
\begin{itemize}
	\item Relevance: Whether the explanation adheres to the specific safety rule.
	\item Equivalence: Whether the explanation is talking about the same violation, in terms of object, and reason, as the ground truth.
	\item Specificity: Whether the explanation pinpoints the specific violator by describing the location or attribute. 
\end{itemize}

Drawing inspiration from \citep{Liu2023llava,Hodosh2013Flickr8k,Zheng2023judgingllmasajudge} and to expedite the evaluation process, we employ another LLM, Meta Llama 3 8B Instruct \citep{Llama3modelcard}, as a judge. The judge assigns three integer marks ranging from 0 to 2 for each of three criteria (i.e., relevance, equivalence, and specificity): 0 indicates incapability for the criterion, 1 suggests attempting but not succeeding well, and 2 signifies acceptability for the criterion. Therefore, the maximum total score across three criteria is 6. For a meaningful evaluation, we only assess reasoning and visual grounding for correctly selected violations. 

Evaluation of reasoning for each rule was conducted separately. We provided the judge with three human judgment examples from the for the evaluation of each rule to create a three-shot setting \citep{Zheng2023judgingllmasajudge}, and asked it to evaluate all the VLM reasonings for each rule. Spearman's rank correlation test and Pearson correlation test were employed to measure the correlations between human evaluation and judge evaluation \citep{Machacek2014WMTMetrics}. A resulting Spearman's coefficient $\rho = 0.83$, and a Pearson's coefficient $r = 0.91$ proves a strong monotonic and linear relationship between human evaluation and judge evaluation. To ensure consistent and deterministic model behavior, we utilize the beam search method with a fixed random seed of 20, set the number of beams to 5, and consistently use the same three-shot examples for each rule. The beam search method generates a sentence word by word, resembling a tree structure. At each step, a fixed number $k$ (5 in our case) of candidate sentences (beams) are considered when selecting the next word, and the new $k$ candidate beams are retained for the following word.

\subsection*{Visual grounding}
\label{Visual grounding}

Each model will be tasked with grounding three types of objects: excavators, rebars, and workers with white hard hats. These tasks will be presented to the models individually. The difficulty of these tasks increases progressively, because:

\begin{itemize}
	\item Excavators are relatively large and have distinct features.
	\item Rebars vary in shape and can be easily confused with other construction materials.
	\item Workers with white hard hats restrict the detection target to a specific type of worker.
\end{itemize}

This progression in difficulty aims to challenge the models' capabilities in visual grounding across different object types.

The evaluation metric is the IoU of reference and candidate bounding boxes, which is explained in Section \ref{benchmarking: safety rule violation VQA}, and the equation is shown in Eqn. \ref{iou}. The overall workflow mirrors the visual grounding branch depicted in Figure \ref{fig:safety rule violation vqa workflow}.

\section{Results and Discussion}
\label{results and discussion}

\subsection*{Resource usage}

Experiments for open-source VLMs are conducted on NVIDIA RTX4080 (16GB) GPU or NVIDIA Tesla V100 High Performance Computer. Proprietary models are running with respective official API with 900 Mbps download and 200 Mbps upload internet connection. Due to GPU memory constraint, open-source VLMs with more than 7B parameters need to be 4-bit quantized when running on RTX4080. We summarized the resource usage in Table \ref{tab:resource usage}. As the resource usage differences within large proprietary models and within smaller open-source models are minimal, we report results only for Qwen-2.5-VL and Gemini-2.5-Flash, which serve as up-to-date representative examples for 2025. While larger proprietary models provide stronger reasoning ability, smaller open-source VLMs offer a favorable trade-off between accuracy and scalability, making them more suitable for resource-constrained industrial adoption.

\begin{table*}[htb!]
	\centering
	\caption{The reported resource usage reflects the consumption per image after completing one round of image captioning (5-shot), VQA (5-shot), and object grounding (for three object as per our test). The cost estimates are derived from the total number of tokens processed, with the note that proprietary models typically apply different pricing schemes depending on the type of token.}
	\label{tab:resource usage}
	\resizebox{\textwidth}{!}{%
		\begin{tabular}{lcccccc}
			\hline
			& \multicolumn{3}{c}{Inference latency per image (seconds)} & \multicolumn{3}{c}{Average resource consumption per image} \\ \cline{2-7} 
			& Image captioning (5-shot)   & \makecell[c]{VQA \\ (5-shot for proprietary models)}   & Object grounding   & Token usage   & Cost (USD)   & GPU usage (seconds)  \\ \cline{2-7} 
			Proprietary models & 1.7                    & 2.1      & 1.9                  &  5651             &  0.0015              &  -     \\
			Open-source models               &  4.5                   & 1.9      & 0.8                  &  -             &  -              & 8.8      \\ \hline
		\end{tabular}%
}
\end{table*}

\subsection*{Image captioning}

\begin{figure*}[htb!]
	\centerline{\includegraphics[width=\textwidth, height=0.85\textheight]{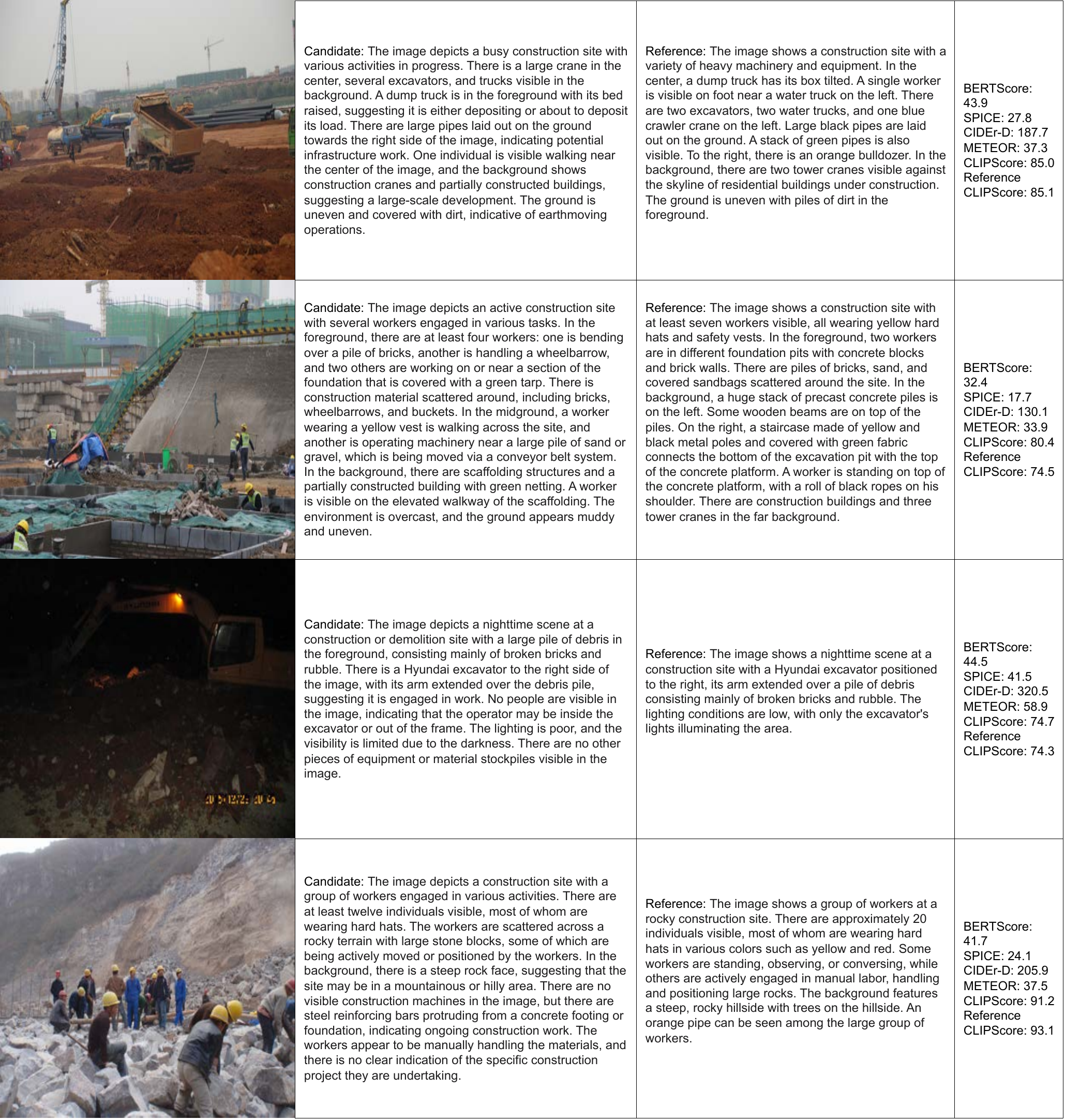}}
	\caption{The figure displays examples of candidate captions generated by the GPT-4 model, alongside reference captions and their evaluations (all scores are in \%) for the image captioning task. For all evaluation metrics except the one assessing the reference caption, higher scores indicate a ``better" candidate caption. While the five types of image captioning metrics assign different scores to the same caption, they generally exhibit a monotonic relationship. This means that if one metric assigns a higher score to a specific caption, it is likely that the others will also assign higher scores.}
	\label{fig:image captioning example}
\end{figure*}

Examples of the image captioning task is presented in Figure \ref{fig:image captioning example}. The result for the image captioning task is summarized in Table \ref{tab:image caption result}. For the three rule-based metrics (i.e. SPICE, CIDEr-D, METEOR), we present the SOTA scores from the MSCOCO dataset benchmarks for comparison, while for the CLIPScore, we present a human baseline. It should be noted that MSCOCO SOTA provide insight into the performance of models pretrained or fine-tuned with the MSCOCO dataset, making this a slightly unfavorable comparison for models trained using the \textit{ConstructionSite 10k} dataset. On one hand, captions in MSCOCO typically contain significantly fewer words on average compared to those in \textit{ConstructionSite 10k}. On the other hand, images in the \textit{ConstructionSite 10k} dataset include more semantic complexity, making it more challenging for models to capture all details accurately. 

The results demonstrate that powerful models like Gemini-2.5-VL and GPT4o, tested under a 5-shot setting, achieve comparable or higher scores than the SOTA benchmarks from MSCOCO for the three rule-based metrics. Even without few-shot examples, the GPT4V model shows a strong ability to understand construction site images and effectively express its understanding, outperforming LLaVA models.

The outstanding performance of GPT4V 5-shot compared to the zero-shot setting demonstrates the undeniable in-context learning capacity of GPT models. Despite no changes to the image, text encoders, or decoder, these VLMs achieve higher scores by aligning their captions closely with human-annotated versions in terms of terminology, length, and format. This suggests that large pre-trained VLMs can quickly adapt to new tasks with minimal demonstrations. 

In the \textit{ConstructionSite 10k} dataset, Gemini model set a new SOTA record in the image captioning task. MiniGPT-4, LLaVA and Qwen models, constrained by their smaller size and less extensive training, do not achieve comparable results to large proprietary models across any metric. Additionally, the LLaVA 13B model does not benefit significantly from the few-shot setting, possibly due to the lack of prompt images.

The CLIPScore, a model-based metric, offers the opportunity to compare image captions directly with images without relying on an intermediary such as a human-written caption. When evaluating images without any reference, all models surprisingly received high scores compared to human annotations. This suggests that all models generate high-quality image captions that exhibit high cosine similarities with the images. However, the limitations of text encoders in encoding long sentences, spatial relationships between objects, and numerical details \citep{Hessel2023textencoderbottleneck}, coupled with potential gaps in training data related to construction sites, may affect the fidelity of CLIPScore.

\begin{table*}[htb!]
	\centering
	\caption{The evaluation results (in \%) for image description with automatic metrics. Higher scores in SPICE, CIDEr-D, METEOR, BERTScore indicate greater similarity between human-annotated and VLM-generated descriptions. In CLIPScore, higher scores indicate greater similarity between machine-generated descriptions and images. MSCOCO SOTA scores for the three rule-based metrics are included to provide context but are not directly comparable with other scores in this table. The best-performing VLM results are underlined.}
	\label{tab:image caption result}
	\resizebox{\textwidth}{!}{%
		\begin{tabular}{lcccccc}
			\hline
			Model            & SPICE & CIDEr-D & METEOR & BERTScore & CLIPScore & \multicolumn{1}{l}{Average words per caption} \\ \hline
			GPT4V            & 18.2  & 120.5   & 33.7   & 28.0      & \underline{82.2}      & 119                                           \\
			GPT4V 5-shot     & 23.1  & 152.1   & 38.7   & 33.5      & 82.1      & 101                                           \\
			GPT4o 5-shot     & 24.9  & 169.9   & \underline{39.4}   & 37.7      & 81.8      & 80                                            \\
			Gemini-2.5 5-shot     & \underline{26.2}  & \underline{183.0}   & 38.6   & \underline{38.9}      & 81.4     &  69                                          \\
			MiniGPT4         & 15.5  & 73.2    & 28.3   & 31.1      & 77.9      & 60                                            \\
			LLaVA 7B         & 12.8  & 62.1    & 28.5   & 26.4      & 78.6      & 82                                            \\
			LLaVA 13B        & 14.6  & 69.2    & 29.8   & 27.8      & 79.6      & 84                                            \\
			LLaVA 13B 5-shot & 15.1  & 70.5    & 30.8   & 27.2      & 76.5      & 101                                           \\ 
			Qwen 7B 5-shot & 20.2  & 119.4   & 32.9   & 33.1      & 82.1      & 72                                           \\ \hline
			Average & 19.0     & 113.3       & 33.4     & 31.5         & 80.2      & 85
			\\
			Median & 18.2     & 119.4       & 32.9     & 31.1         & 81.4      & 82
			\\
			Human annotation & -     & -       & -      & -         & 80.0      & 49                                            \\
			MSCOCO SOTA\footnotemark      & 27.0  & 155.1   & 33.9   & -         & -         & -                                             \\ \hline
		\end{tabular}%
}
\end{table*}
\footnotetext{MSCOCO SOTA results are obtained from website: https://paperswithcode.com/sota/image-captioning-on-coco-captions}

VLMs tend to generate verbose descriptions even when prompted to be concise. The word count analysis reveals that models typically use twice as many words to describe an image compared to human annotators. Interestingly, GPT4o achieves the best overall performance with the fewest words among the models. This phenomenon may stem from harsh penalties imposed on candidate sentences that significantly exceed the length of reference captions.

\subsection*{Safety rule violation VQA}
\label{result: safety rule violation VQA}

Before evaluation, we conduct post-processing on candidate captions to ensure they are readable, free of syntax errors, and uniformly formatted for fair automatic evaluation. We did not include testing results of MiniGPT-4 and LLaVA 7B due to their inconsistent ability to generate useful responses in VQA and object grounding tasks. Similarly, smaller open-source VLMs like LLaVA 13B and Qwen 7B with around 10B parameters were not tested in few-shot scenarios, as they struggle to leverage in-context learning and sometimes repeat examples in their answers.

Additionally, we conducted the same test with human participants who are graduate students in civil engineering, who received the exact same image-prompt pairs as the VLMs.

\begin{figure*}[htb!]
	\centerline{\includegraphics[width=\textwidth]{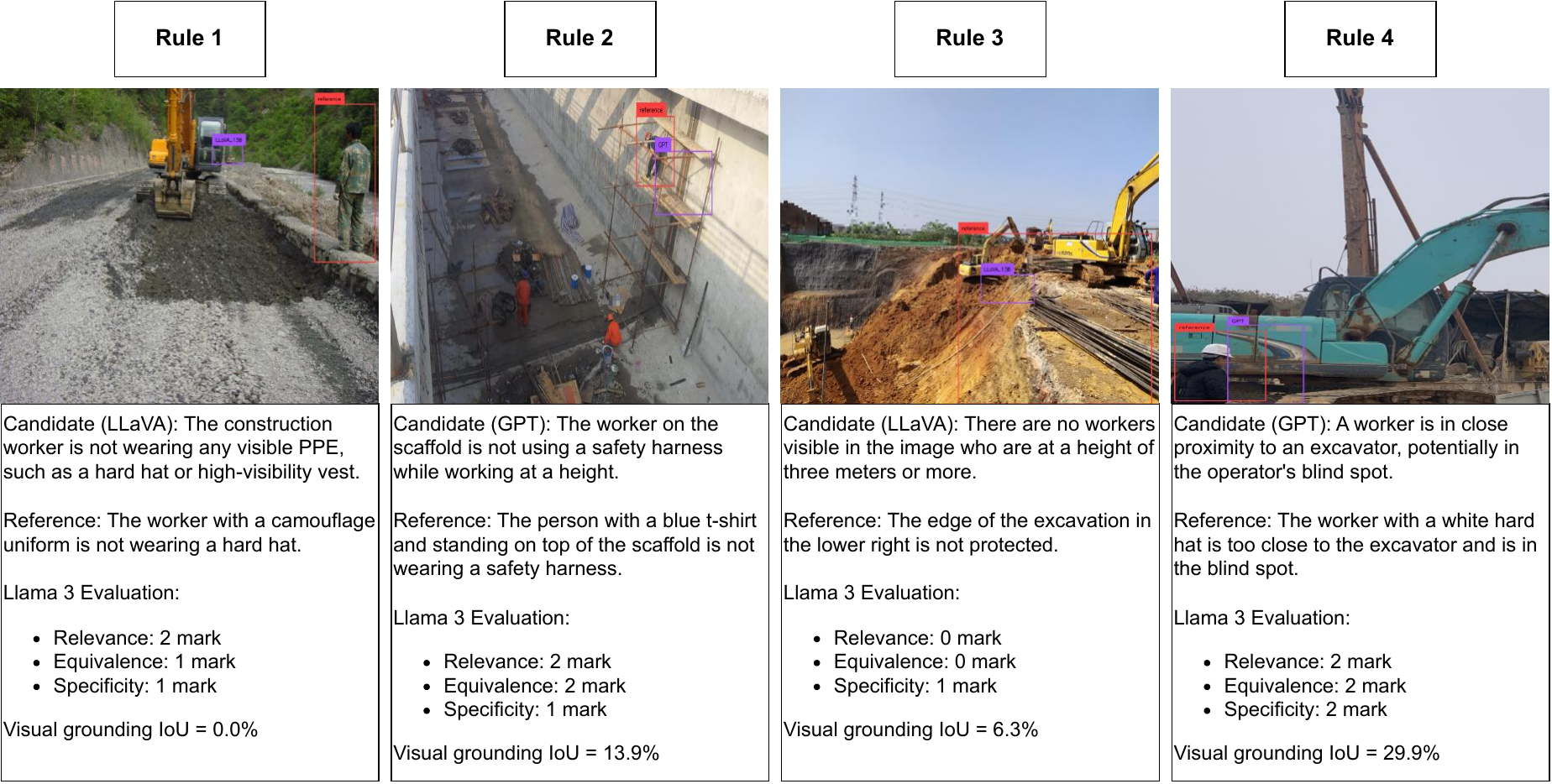}}
	\caption{Examples of safety rule violation reasoning. The figure includes the candidate, reference reasoning, and the evaluations.}
	\label{fig:examples of vqa reasoning with no comment}
\end{figure*}

The testing results are presented in Table \ref{tab:multilabel classification results} and Table \ref{tab:safety rule violation reasoning results}. To provide a clearer visualization of the results and evaluations, examples of safety rule violation reasoning are presented in Figure \ref{fig:examples of vqa reasoning with no comment}.

High precision scores across all rules for human performance validate the ground truth. However, relatively lower recall rates indicate that construction engineers may overlook some dangers at construction sites, even when violations of safety protocols are obvious. 

For the violation detection, the high recall rates and significantly lower precision of all VLMs suggest that the models tend to be conservative and may overshoot in identifying safety violations. Additionally, the results indicate that the chain-of-thought method does not significantly improve model performance when tests do not involve strong sequential dependencies, as demonstrated in previous studies \citep{Liu2023llava,Hessel2023NewYorkcaption}. A closer examination of numbers reveals that the models perform relatively worse on safety rules 2 and 4 compared to human upper bounds. Safety rule 2 requires workers working at heights to wear safety harnesses, while safety rule 4 requires workers to stay clear of an excavator’s operating radius. Both rules demand an understanding of spatial relationships within the image. Specifically, safety rule 2 requires the VLM to determine whether a worker is working at height, whereas safety rule 4 requires assessing whether a person is near the excavator and reasoning about the potential risk of being hit. Humans—especially expert civil engineers—can easily arrive at the correct conclusion using spatial reasoning and common sense. The results, however, indicate that the VLM still struggles with these types of spatial and logical reasoning tasks.

In terms of reasoning, only GPT and Gemini models approach the upper bounds set by human-level textual explanations. Open-source LLaVA and Qwen models, utilizing smaller language models, struggle to produce high-quality textual explanations. Regarding visual grounding, none of the models accurately pinpoint the exact location of violations, as evidenced by notably low IoU scores. The low IoU scores further reflect the limited spatial reasoning capability discussed above, particularly for the LLaVA and Gemini models, which are either smaller in model size or are not explicitly reported to be trained on object localization or bounding box generation tasks. These findings are visualized in Figure \ref{fig:examples of vqa reasoning with no comment}. In the worst cases, some models even fail to provide valid bounding box coordinates for evaluation.

Contrary to the intuition that ``larger models perform better on everything", Qwen 7B demonstrates impressive visual grounding capacity in safety rule violation detection. This behavior may stem from Qwen’s pre-training strategy, which uses absolute pixel coordinates for object localization \citep{Bai2025Qwen2.5vl}. This approach likely enhances its visual grounding capability compared to models trained on normalized coordinates, as is common in visual instruction tuning \citep{Liu2023llava}. We blame the poor grounding capacity of large proprietary models- Gemini and GPT series- on the absence of specific visual grounding training \citep{Comanici2025Gemini2.5,Openai2024gpt4}. 

We also acknowledge the limitations of the safety rule violation VQA:
\begin{itemize}
	\item The four safety rules do not exhaustively cover all potential safety violations at construction sites.
	\item While humans can readily estimate distances in 2D images using visual comparisons and common sense, quantitative estimations (e.g., “three meters” in our work) remain challenging for VLMs, as their training data often lack precise and reliable spatial annotations.
	\item Localizing safety rule violations using bounding boxes may be imprecise, since a VLM can achieve substantial overlap with the ground-truth bounding box while still failing to accurately capture the true region or underlying nature of the violation. In particular, for geometrically complex cases—such as inclined or partially occluded workers—bounding boxes may overshoot the relevant area. More precise annotation formats, such as segmentation masks, could serve as a promising alternative in future work.
\end{itemize}

\begin{table*}[htb!]
	\caption{The result (in \%) of the safety rule violation detection. Rule 0 indicates no violation and therefore cannot co-exist with other rules, whereas other rules can co-exist within the same image. The models answer the question in a multi-label classification setting. The best-performing VLM results are underlined.}
	\label{tab:multilabel classification results}
	\resizebox{\textwidth}{!}{%
		\begin{tabular}{lcccccccccc}
			\hline
			Rule ID & \multicolumn{2}{c}{0} & \multicolumn{2}{c}{1} & \multicolumn{2}{c}{2} & \multicolumn{2}{c}{3} & \multicolumn{2}{c}{4} \\ \hline
			& \multicolumn{1}{l}{precision} & \multicolumn{1}{l}{recall} & \multicolumn{1}{l}{precision} & \multicolumn{1}{l}{recall} & \multicolumn{1}{l}{precision} & \multicolumn{1}{l}{recall} & \multicolumn{1}{l}{precision} & \multicolumn{1}{l}{recall} & \multicolumn{1}{l}{precision} & \multicolumn{1}{l}{recall} \\ \cline{2-11} 
			GPT4V & 97.0 & 24.6 & 20.4 & 76.4 & 2.7 & \underline{85.7} & 5.3 & 87.7 & 3.7 & \underline{84.0} \\
			GPT4V 5-shot & \underline{98.9} & 32.0 & 18.2 & \underline{89.4} & 5.4 & \underline{85.7} & 4.3 & \underline{89.2} & 4.4 & 68.0 \\
			Gemini-2.5 5-shot & 97.5 & 62.1 & \underline{25.6} & 81.7 & \underline{8.8} & 76.0 & \underline{7.0} & 39.7 & 5.0 & 70.8 \\
			LLaVA 13B & 88.0 & 43.4 & 12.0 & 54.0 & 1.0 & 42.9 & 2.2 & 33.8 & 0.9 & 44.0 \\
			LLaVA 13B CoT & 91.0 & 44.0 & 12.0 & 55.0 & 3.0 & 52.0 & 6.0 & 18.0 & 2.0 & 32.0 \\
			LLaVA 34B 1-shot & 87.1 & \underline{88.7} & 14.3 & 13.0 & 3.0 & 33.3 & 4.5 & 21.5 & \underline{5.2} & 20.0 \\ 
			Qwen 7B & 91.2 & 42.3 & 12.9 & 71.5 & 2.3 & 48.0 & 3.4 & 39.7 & 1.7 & 54.2 \\\hline
			Average & 93.0 & 48.2 & 16.5 & 63.0 & 3.7 & 60.5 & 4.7 & 47.1 & 3.3 & 53.3 \\
			Median & 91.2 & 43.4 & 14.3 & 71.5 & 3.0 & 52.0 & 4.5 & 39.7 & 3.7 & 54.2 \\
			Human upper bound & 95.3 & 98.3 & 95.6 & 66.6 & 92.0 & 92.0 & 59.4 & 60.3 & 81.0 & 65.4 \\ \hline
		\end{tabular}%
	}
\end{table*}

\begin{table*}[htb!]
	\caption{The table presents reasoning results for safety rule violation VQA. “Violations detected” denotes the number of images in which the respective safety rule violations were correctly identified. “Average score from LLM Judge” ranges from 0 to 6. Intersection over Union (IoU) is reported as a percentage (\%). The best-performing VLM results are underlined.}
	\label{tab:safety rule violation reasoning results}
	\tiny
	\resizebox{\textwidth}{!}{%
	\begin{tabular}{lcccccc}
		\hline
		Rule ID & \multicolumn{3}{c}{1} & \multicolumn{3}{c}{2} \\ \hline
		& \makecell[c]{violations detected \\ (out of 323)} & \makecell[c]{average score \\ from LLM judge} & IoU & \makecell[c]{violations detected \\ (out of 25)} & \makecell[c]{average score \\ from LLM judge} & IoU \\ \cline{2-7} 
		GPT4V & 246 & 4.5 & 11.9 & 18 & 5.7 & 12.1 \\
		GPT4V 5-shot & \underline{288} & 4.7 & 14.0 & 18 & 5.7 & 13.1 \\
		Gemini-2.5 5-shot & 264  & \underline{5.0}  & 3.7 & \underline{19} & 5.7  & 2.5 \\
		LLaVA 13B & 174 & 2.9 & 20.0 & 9 & 5.0 & 17.9 \\
		LLaVA 13B CoT & 177 & 3.1 & 6.9 & 11 & 1.0 & 10.7 \\
		LLaVA 34B 1-shot & 42 & 3.9 & - & 7 & \underline{5.9} & - \\ 
		Qwen 7B & 231 & 3.1 & \underline{23.1} & 12 & \underline{5.9} & \underline{40.1} \\\hline
		Average & 203 & 3.9 & 13.3 & 13 & 5.0 & 16.1 \\
		Median & 231 & 3.9 & 13.0 & 12 & 5.7 & 12.6 \\
		Human upper bound & 215 & 5.2 & - & 23 & 5.9 & - \\ \hline
		& \multicolumn{3}{c}{3} & \multicolumn{3}{c}{4} \\ \cline{2-7} 
		& \makecell[c]{violations detected \\ (out of 63)} & \makecell[c]{average score \\ from LLM judge} & IoU & \makecell[c]{violations detected \\ (out of 24)} & \makecell[c]{average score \\ from LLM judge} & IoU \\ \cline{2-7} 
		- & 57 & 5.6 & 30.6 & \underline{21} & 5.7 & 10.0 \\
		- & \underline{58} & \underline{5.7} & \underline{32.6} & 17 & \underline{5.9} & 22.5 \\
		- & 25 & \underline{5.7} & 4.9 & 17 & \underline{5.9} & 9.4 \\
		- & 22 & 3.7 & 8.3 & 11 & 4.6 & 21.5 \\
		- & 12 & 4.8 & 0.5 & 8 & 1.1 & 0.0 \\
		- & 14 & 5.3 & - & 5 & 5.6 & - \\ 
		- & 25 & 5.1 & 25.0 & 13
		 & 5.8 & \underline{25.8} \\ \hline
		- & 30 & 5.1 & 17.0 & 13 & 5.0 & 14.9 \\
		- & 25 & 5.3 & 16.7 & 13 & 5.7 & 15.8 \\
		- & 38 & 5.8 & - & 17 & 6.0 & - \\ \hline
		\end{tabular}%
	}
\end{table*}

\subsection*{Visual grounding}

The result for visual grounding are presented in Table \ref{tab:visual grounding result concatenated}. The results consist of two sections. One section shows the result of all images, no matter the object to ground exist in the image or not, which means false positive cases (no object in the image but bounding box coordinates are given) and false negative cases (object presents in the image but bounding box coordinates are not given) will get an IoU of 0, we call it ``IoU-Total". The other section only account for cases where the object to ground exists in the image, which we call ``IoU-Object Exist". The equations showing the two types of IoU are shown in Eq. \ref{iou total} and Eq. \ref{iou object exist}.

\begin{equation}
        \label{iou total}
		IoU-Object\; Exist = \frac{intersection(\hat{X}, X\, |\, 1\in X)}{union(\hat{X}, X\, |\, 1\in X)}
\end{equation}

\begin{equation}
	\label{iou object exist}
	IoU-Total = \frac{intersection(\hat{X}, X)}{union(\hat{X}, X)} 
\end{equation}

Where $\hat{X}$ is the candidate binary map and $X$ is the reference binary map. ``IoU-Object Exist" demonstrates the ability of VLMs to precisely locate an object in an image by outputting bounding box coordinates, while ``IoU-Total" also reflects their capability to determine whether the object exists. We will primarily examine ``IoU-Object Exist" as our focus is on evaluating the current precision of the VLMs. 

\begin{table*}[hbt!]
    \centering
    \caption{The IoU result (in \%) of object detection results. The ``IoU-Total" result is calculated across all images, while the ``IoU-Object Exist" result is calculated on images where the object to ground exists. Macro averaging refers to calculating IoU for individual images first and then taking the average, micro averaging refers to dividing the summed intersection by the summed union. The macro averaging treats each image equally, while micro averaging awards correct predictions when the object is larger in the image.}
    \label{tab:visual grounding result concatenated}
    \resizebox{\textwidth}{!}{%
        \begin{tabular}{lcccccc}
	        \hline
	         & \multicolumn{2}{c}{Excavators} & \multicolumn{2}{c}{Rebars} & \multicolumn{2}{c}{\begin{tabular}[c]{@{}c@{}}Workers with \\ white hard hats\end{tabular}} \\ \hline
	         & \multicolumn{1}{l}{macro-averaged} & \multicolumn{1}{l}{micro-averaged} & \multicolumn{1}{l}{macro-averaged} & \multicolumn{1}{l}{micro-averaged} & \multicolumn{1}{l}{macro-averaged} & \multicolumn{1}{l}{micro-averaged} \\ \hline
	        \multicolumn{7}{l}{IoU-Object Exist} \\ \hline
	        GPT4V & 28.6 & 35.8 & 10.9 & 18.2 & 5.9 & 10.1 \\
	        Gemini-2.5 & 60.8 & 60.0 & 3.2 & 4.8 & 8.6 & 5.9 \\
	        LLaVA 7B & 31.9 & 34.7 & 8.4 & 13.4 & 14.0 & 15.0 \\
	        LLaVA 13B & 47.7 & 54.5 & 14.9 & \underline{19.0} & 19.5 & 25.4 \\ 
 	        Qwen 7B & \underline{64.7} & \underline{75.4} & 4.0 & 10.7 & \underline{32.1} & \underline{31.5} \\
	        Grounding DINO & 63.9 & 71.0 & \underline{15.6} & 15.3 & 12.9 & 11.2 \\ \hline
	        Average & 49.6 & 55.2 & 9.5 & 13.6 & 15.5 & 16.5 \\
	        Median & 54.3 & 57.3 & 9.7 & 14.4 & 13.5 & 13.1 \\
	        \makecell[l]{Traditional CV model, \\ Confidence threshold = 0.50}  & 65.9 & 67.3 & 45.5 & 50.1 & 41.5 & 37.9 \\
	        \makecell[l]{Traditional CV model, \\ Confidence threshold = 0.75}  & 69.9 & 69.7 & 34.6 & 32.4 & 37.8 & 32.6 \\ \hline \\ \hline
	        \multicolumn{7}{l}{IoU-Total} \\ \hline
	        GPT4V & 22.3 & 29.3 & \underline{5.7} & \underline{11.6} & 3.1 & 9.1 \\
	        Gemini-2.5 & 55.7 & 58.7 & 2.0 & 2.6 & 5.4 & 5.6 \\
	        LLaVA 7B & 11.5 & 21.0 & 1.0 & 3.1 & 1.5 & 3.8 \\
	        LLaVA 13B & 17.4 & 32.6 & 1.7 & 3.8 & 2.1 & 3.2 \\ 
	        Qwen 7B & \underline{56.5} & \underline{67.2} & 3.5 & 7.7 & \underline{11.6} & \underline{21.4} \\
	        Grounding DINO & 42.8 & 45.4 & 2.6 & 2.0 & 3.0 & 2.1 \\ \hline
	        Average & 34.4 & 42.4 & 2.7 & 5.1 & 4.4 & 7.5 \\
	        Median & 32.6 & 39.0 & 2.3 & 3.5 & 3.0 & 4.7 \\ 
	        \makecell[l]{Traditional CV model, \\ Confidence threshold = 0.50}  & 51.1 & 53.9 & 27.5 & 43.5 & 18.4 & 25.1 \\
	       \makecell[l]{Traditional CV model, \\ Confidence threshold = 0.75}  & 62.8 & 63.5 & 28.2 & 31.1 & 26.6 & 27.0 \\ \hline
        \end{tabular}%
    }
\end{table*}

In the object visual grounding task, LLaVA and Qwen models also show comparable or higher performance than the GPT and Gemini model, which echoes the analysis of more spatial training data mentioned in Section \ref{result: safety rule violation VQA}. The lower accuracy of smaller open-source models on rebars is likely due to their limited exposure to such objects during pre-training; rebars are relatively uncommon in general image datasets and are predominantly found in construction-specific environments.

A closer examination of the results reveals clear trends. For unconstrained object categories such as excavators and rebars, earlier generative VLMs like GPT-4V and LLaVA perform substantially worse than Grounding DINO, which is optimized for visual grounding. However, Grounding DINO’s advantage diminishes when the task requires grounding more specific object types, where its performance begins to deteriorate. Notably, newer VLMs released in 2025—such as Gemini-2.5-Flash and Qwen-2.5-VL—show substantial improvements over their predecessors, with Qwen-2.5-VL (7B) achieving the best performance in our visual grounding evaluation. We also noticed that in most cases, the micro-averaged IoU equals or exceeds the macro-averaged IoU, indicating that VLMs typically achieve better results when objects are large in the images.

Overall, due to inferior performance in grounding rebars and workers wearing white hard hats, there remains substantial room for improvement for generative VLMs. Presently, generative VLMs cannot reliably function as visual grounding models at construction sites, particularly under realistic conditions where tasks impose additional constraints on attributes.

Examining the ``IoU-Total" section reveals that when false positive cases are considered, the performance of LLaVA and Qwen models is significantly impacted. Specifically, LLaVA's performance in grounding excavators is halved, and there is a 90\% decrease in grounding rebars and workers with hard hats compared to ``IoU-Object Exist." Qwen also experienced a 60\% drop in detecting workers with white hard hats. Since in the ``IoU-Object Exist" cases, we assist the models in eliminating false positive predictions, this difference in results indicates the two open-source models tend to make bold guesses, resulting in the accidental capture of a substantial portion of the object. Thus, the high IoU rates observed for the LLaVA and Qwen models may be attributed to chance predictions, as they tend to provide predictions for all images. Conversely, the modest decline in the performance of GPT and Gemini demonstrates its cautious approach in making decisions.

For this task, we include a Faster R-CNN model as a baseline to compare the out-of-context grounding capabilities of large pre-trained VLMs—which are stronger in in-context text-based reasoning—with a CNN-based model specialized in object detection. We report IoU-based performance at two confidence thresholds: 0.5 and 0.75, meaning a predicted bounding box is considered valid if its confidence score exceeds the threshold. The 0.5 threshold provides an overall view of coverage, while the 0.75 threshold emphasizes higher-confidence predictions with more precise localization. As shown in Table \ref{tab:visual grounding result concatenated}, the traditional CV model achieves IoU scores approximately twice the average VLM IoU for excavator detection, and several times higher for rebars and workers with white hard hats, demonstrating significantly better bounding box grounding capability. However, this comparison may be somewhat biased due to the availability of task-specific training data for the Faster R-CNN model, while all large pre-trained VLMs are not fine-tuned on task-specific data. The traditional CV model’s limited generalization to new tasks and unseen object types highlights the trade-off between specialized detection and the multi-modal reasoning capabilities of VLMs in construction site safety inspection.

\begin{figure*}[htb!]
	\centerline{\includegraphics[width=0.75\textwidth]{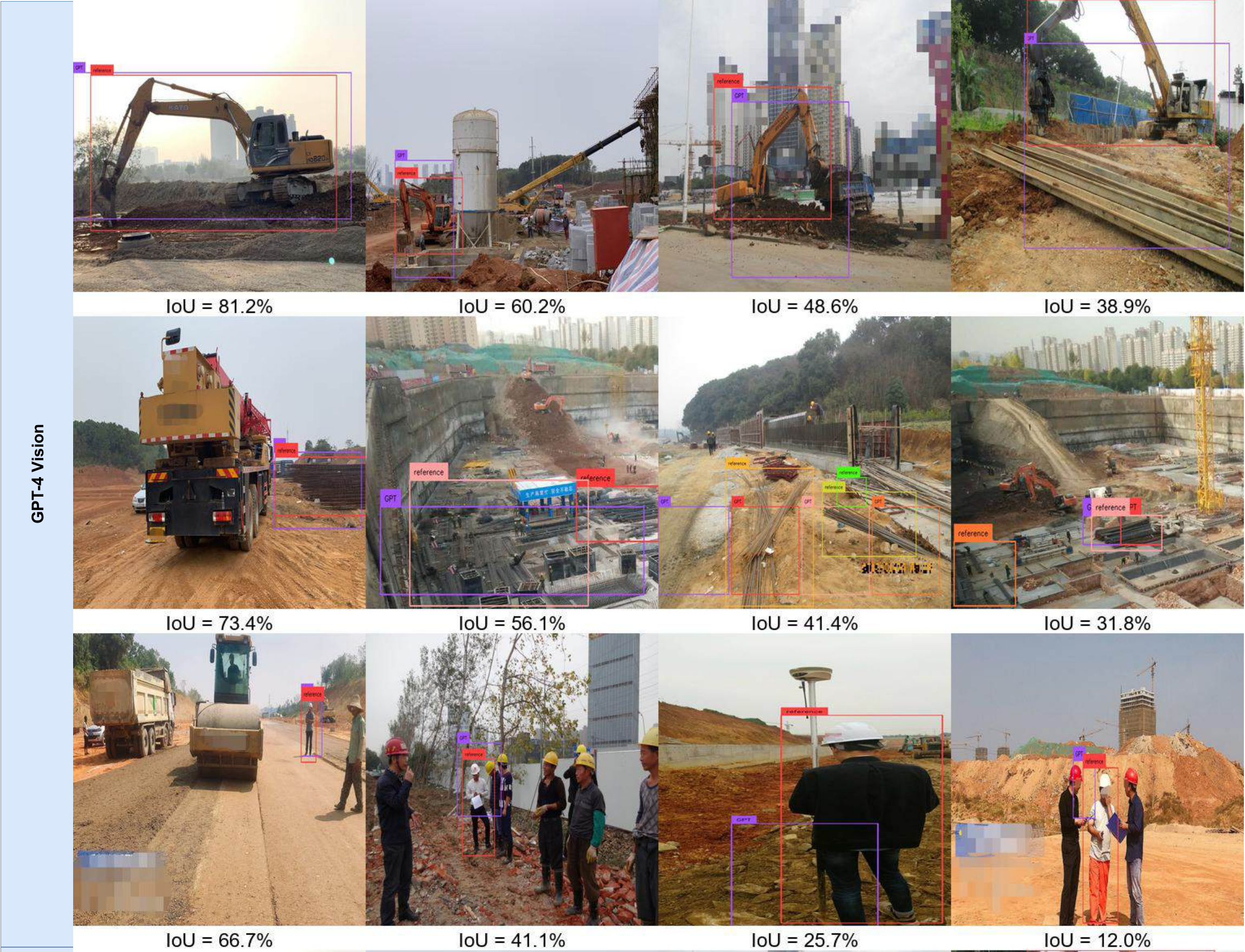}}
	\caption{The image displays visual grounding examples from GPT model. Each row corresponds to a different object category: the first row shows excavators, the second row shows rebar detection, and the third row shows workers with white hard hats. The model excels in detecting larger objects but struggles with irregular shapes, such as rebar piles, and specific constraints, such as workers with white hard hats.}
	\label{fig:detection examples}
\end{figure*}

To better visualize the performance, we presents some visual grounding results in Figure \ref{fig:detection examples}.

\subsection*{Implications for VLM Design and Selection in Construction Safety Inspection}

The evaluated models exhibit substantial performance variation across different tasks. Across experiments, we observe consistent distinctions between (i) large proprietary, instruction-tuned VLMs and (ii) open-source VLMs constructed by aligning vision encoders with large language models, with Qwen-2.5-VL as a notable exception.

Models with extensive multimodal pre-training and strong instruction tuning consistently perform better on language-intensive tasks. This trend is evidenced by the generally above-average and above-median performance of GPT and Gemini models in image captioning and safety rule violation detection and reasoning. Their strong performance suggests that semantic alignment, instruction-following capability, and robustness to long and complex prompts are critical for high-level safety understanding, coherent textual reasoning, and consistent output formatting. While these proprietary models are closed-source and their full architectural details are unavailable, their reported designs \cite{Openai2024gpt4, Comanici2025Gemini2.5} indicate that large model capacity, transformer-based vision–language encoders, massive multimodal pre-training, and subsequent instruction tuning jointly contribute to superior language-centric performance.

In contrast, spatial grounding tasks—such as localizing safety rule violations—remain challenging for all evaluated VLMs. Average and median IoU scores are generally below 20\%, with the exception of grounding excavators, which are visually salient objects with relatively simple geometry. Notably, Qwen-2.5-VL, which explicitly incorporates detection-oriented pre-training with accurate coordinates \cite{Bai2025Qwen2.5vl}, consistently outperforms other open-source models in IoU metrics and achieves grounding performance comparable to GPT-4V and Gemini-2.5. This observation indicates that strong language reasoning alone does not guarantee accurate visual grounding. Instead, targeted spatial supervision can substantially enhance grounding performance, enabling smaller open-source models to match or exceed the grounding capability of much larger proprietary models.

These findings lead to a practical question for construction safety inspectors and researchers: \textit{Which type of VLM is most suitable for a given inspection task?} Since all evaluated VLMs are transformer-based, differences in task performance are better explained by model scale, pre-training objectives, and the inclusion of spatial supervision rather than by architectural variations within the transformer family. Our results suggest that instruction-tuned multimodal models are preferable for reasoning-heavy, open-ended inspection tasks that require semantic understanding and natural language explanation. Conversely, tasks demanding precise geometric localization benefit more from models trained with explicit visual grounding data. Given the consistently limited grounding accuracy across VLMs, closed-set detection tasks may still be better served by traditional computer vision models, which remain more reliable for narrowly defined, single-purpose detection problems.

As new VLMs continue to emerge, their suitability for specific construction safety inspection scenarios can be anticipated based on their training objectives, degree of spatial supervision, and balance between vision and language capacity. We emphasize that these insights are derived from empirical trends observed in this benchmark rather than from exhaustive architectural ablation studies.

\section{Conclusion}
\label{conclusion}

This paper introduces \textit{ConstructionSite 10k}, a multi-modal dataset comprising 10,013 diverse construction site images and three different tasks designed for pre-training or fine-tuning purposes. The image captioning task provides detailed descriptions of construction scenes, encompassing background details and inter-object relationships. This task aims to train and evaluate VLMs on their understanding of construction site images. The VQA task features annotations identifying safety rule violations along with corresponding bounding boxes indicating the violation locations. The visual grounding task includes bounding boxes for three key construction elements: excavators, rebars, and workers wearing white hard hats. These annotations are intended to train and assess VLMs in tasks related to construction site inspection.

We evaluate the capability of popular off-the-shelf cloud-based and open-source large pre-trained VLMs to generate captions for construction site images without specific construction-related training. Performance assessment employs both rule-based and model-based metrics. In a five-shot setting, the top-performing Gemini model demonstrates superior image comprehension and in-context learning capacity compared to other smaller models.

In the safety violation VQA testing section, we devised a three-stage framework to assess whether models, without supervised training, can understand safety rules, identify violations, justify their answers with reasoning, and accurately draw bounding boxes to localize violation behaviors. We utilized Llama 3 as an evaluator to assess the quality of reasoning. Our findings indicate that for violation detection, most models exhibit higher recall than precision, whereas human experts typically show higher precision than recall. However, nearly all models struggle with accurately grounding violations using bounding boxes. In contrast, in the textual reasoning section, GPT models consistently scored above 5 marks, while LLaVA models can average around 4 marks. 

For object detection, VLMs can perform well for simple objects such as ``excavator" with an IoU above 30\%. But the performance declines notably to below 20\% for objects with irregular shapes or additional constraints, such as workers wearing white hard hats. 

The dataset and testing framework serve as a foundational resource for further training, fine-tuning, and application of large pre-trained VLMs in understanding construction site images and conducting inspection tasks. This infrastructure can effectively benchmark model performance, thereby enhancing model architecture and training strategies. Researchers in the future can benefit from avoiding the manual collection and annotation of datasets, while also being able to benchmark and compare their models against others. This paper pioneers the use of large generative pre-trained VLMs to tackle multiple construction site inspection tasks in zero-shot and few-shot scenarios. It demonstrates the promising capability of these models, even without additional training, to handle construction-related tasks. However, they are still behind their human counterparts in terms of quality and consistency of performance. This research lays the groundwork for the widespread adoption of large pre-trained VLMs or their architectures as the preferred choice for construction inspection tasks.

This paper acknowledges its limitations. We recognize a lack of diversity in the Safety Rule Violation VQA and Construction Element Visual Grounding sections of our dataset. The four safety rules included do not represent all possible safety violations that can occur at construction sites and other visual grounding data types, such as segmentation can be a more promising alternative for geometrically complex cases. Similarly, the number of constrained construction elements in our dataset may not exhaustively cover all types of construction elements. Moreover, we did not fine-tune (i.e., retrain) the large pre-trained models utilized here in our paper. We anticipate that fine-tuning these models could potentially yield superior performance compared to the zero-shot or few-shot approaches employed. Additional fine-tuning on the image captioning task enhances the ability of VLMs to capture the semantic content of images and align them more effectively with corresponding texts. Similarly, fine-tuning on VQA and visual grounding tasks improves the models’ performance on those respective tasks. Additionally, considering the substantial performance gaps between SOTA Gemini model and smaller-scale open-source models like LLaVA or MiniGPT-4, exploring methods to train smaller models to achieve comparable performance to Gemini models hold promise. This direction could prove cost-effective for applications in construction sites, where smaller models may offer practical advantages.

\begin{Backmatter}
	
\section*{Abbreviations}
VLM, vision-language model; VQA, visual question answering; PPE, personal protective equipment; CNN, convolutional neural network; CLIP. Contrastive Language-Image Pre-training; SOTA, state-of-the-art; IoU, intersection over union; CoT, chain-of-thought

\section*{Data Availability Statement}
The dataset is publicly available at: https://huggingface.co/datasets/LouisChen15/ConstructionSite \cite{vlmdataset}.

\section*{Funding Statement}
This work received no specific grant from any funding agency, commercial or not-for-profit sectors.

\section*{Competing Interests}
None

\section*{AI Statement}
This work uses AI tools as part of the research method. Specifically, we benchmark current state-of-the-art AI systems' capabilities to understand safety as a concept in the context of construction. The specific AI models used include the GPT family and LLaVA. 

\clearpage

\bibliographystyle{plainnat}
\bibliography{cas-refs}

\end{Backmatter} 

\end{document}